\documentclass[conference]{IEEEtran}
\newcommand{\eat}[1]{}

\ifCLASSINFOpdf
   \usepackage[pdftex]{graphicx}
\fi

\usepackage[cmex10]{amsmath}

\usepackage{algorithm}
\usepackage{algorithmic}
\usepackage{amsfonts}
\usepackage{tabularx}
\usepackage{amssymb,amsmath}

\usepackage{pbox}
\usepackage{graphicx}
\usepackage[space]{grffile}
\usepackage{epstopdf}
\usepackage{fancyhdr} 

\usepackage{setspace}
\usepackage{graphicx}

\usepackage{comment}
\usepackage{array}
\usepackage{caption}
\usepackage{graphicx}
\usepackage{subcaption}
\usepackage{dblfloatfix}
\usepackage[table]{xcolor}

\usepackage[T1]{fontenc}
\usepackage{newtxtext}
\usepackage{newtxmath}

\usepackage{multirow}

\begin{document}

\title{LLM on a Budget: Active Knowledge Distillation for Efficient Classification of Large Text Corpora}

% Working order 
\author{\IEEEauthorblockN{
Viviana Luccioli, Rithika Iyengar, Ryan Panley, Flora Haberkorn, \\ Xiaoyu Ge, Leland Crane, Nitish Sinha, Seung Jung Lee}
\IEEEauthorblockA{Board of Governors of the Federal Reserve System\\
\{viviana.c.luccioli, rithika.k.iyengar, ryan.p.panley, flora.m.haberkorn, steve.ge, leland.d.crane, nitish.r.sinha, seung.j.lee\}@frb.gov}
\thanks{This work has been submitted to the IEEE for possible publication. Copyright may be transferred without notice, after which this version may no longer be accessible.}
\thanks{The views expressed herein are those of the authors, and do not reflect those of anyone else at the Board of Governors of the Federal Reserve System.}}

\IEEEoverridecommandlockouts

\maketitle

\begin{abstract}
Large Language Models (LLMs) are highly accurate  in classification tasks, however,  substantial computational and financial costs hinder their large-scale deployment in dynamic environments. Knowledge Distillation (KD) where a LLM "teacher" trains a smaller and more efficient "student" model, offers a promising solution to this problem. However, the distillation process itself often remains costly for large datasets, since it requires the teacher to label a vast number of samples while incurring significant token consumption.
To alleviate this challenge, in this work we explore the active learning (AL) as a way to create efficient student models at a fraction of the cost while preserving the LLM's performance. In particular, we introduce M-RARU (Multi-class Randomized Accept/Reject Uncertainty Sampling), a novel AL algorithm that significantly reduces training costs. M-RARU employs an innovative strategy combining uncertainty with a randomized accept-reject mechanism to select only the most informative data points for the LLM teacher. This focused approach significantly minimizes required API calls and data processing time.
We evaluate M-RARU against random sampling across five diverse student models (SVM, LDA, RF, GBDT, and DistilBERT) on multiple benchmark datasets. Experiments demonstrate that our proposed method achieves up to 80\% reduction in sample requirements as compared to random sampling, substantially improving classification accuracy while reducing financial costs and overall training time.
\end{abstract}

%-the actual paper-------------------------------------------------------------

\section{Introduction}
\label{sec:intro}

With the unceasing expansion of unstructured text in the modern data landscape, text classification has become a central tool for extracting insights at scale. For instance, in the financial sector, this capability is especially critical for a diverse array of tasks, ranging from analyzing market trends in news reports and corporate filings to assessing credit risk and ensuring regulatory compliance \cite{loughran2011liability, shapiro2020measuringsentiment}. As the volume and complexity of this textual data grow, a fundamental challenge arises: balancing the trade-off between a model's predictive power and its computational and financial cost. Meeting this challenge is crucial for deploying effective text classification systems in real-world, resource-constrained environments where timely analysis is paramount.

\begin{figure}[t]
        \centering
        \includegraphics[width=0.99\linewidth]{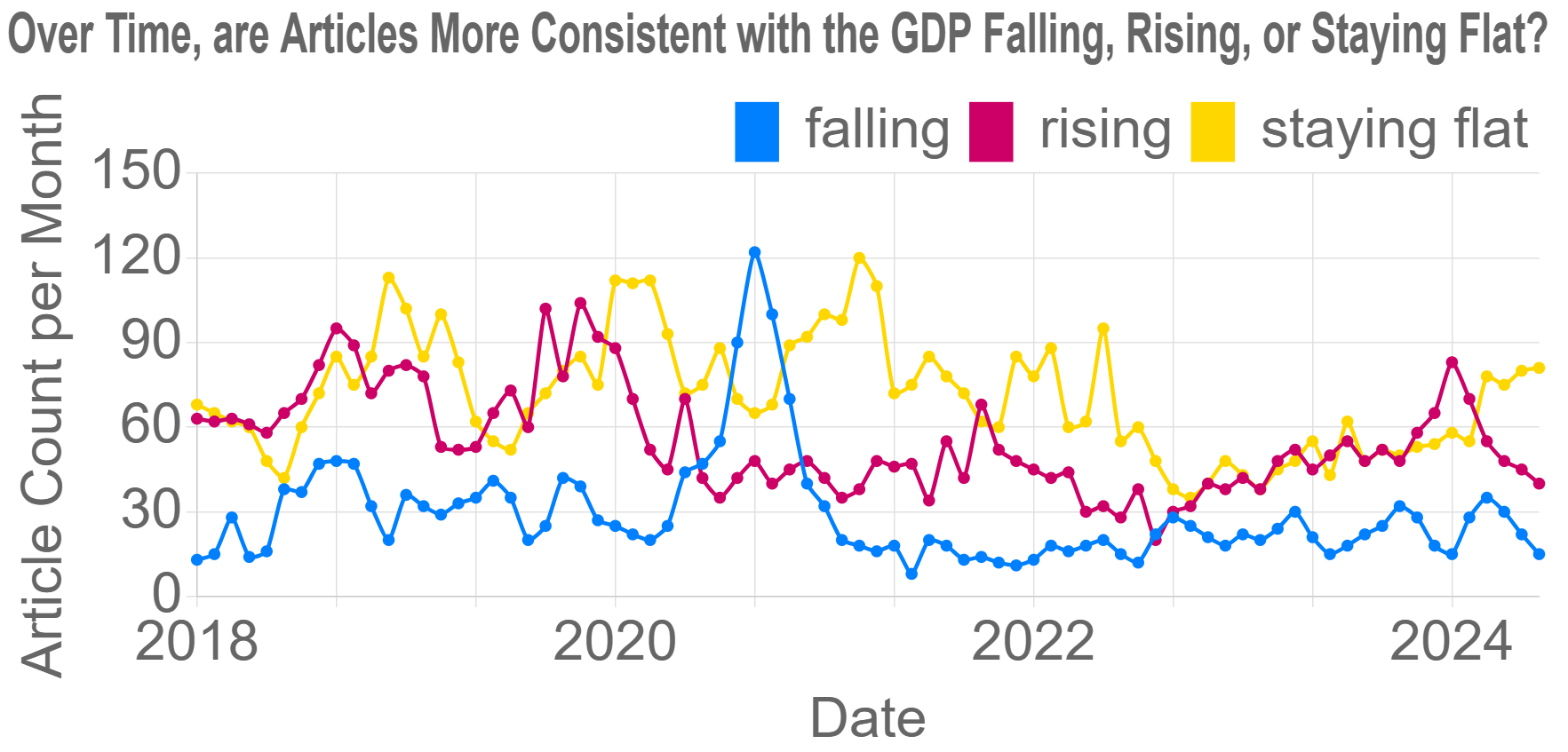} 
        \caption{Text classification for GDP trends.}
        \label{fig:gdpexample} 
        \vspace{-5pt}
\end{figure}

Consider, for example, the task of classifying news articles based on their implications for GDP trends, as illustrated in Figure \ref{fig:gdpexample}. Financial institutions must process thousands of such articles daily to inform investment decisions and economic forecasts. While an LLM can achieve high accuracy in determining whether an article suggests GDP is 'falling,' 'rising,' or 'staying flat,' the computational cost of processing this volume of text at the required speed is prohibitive. Conversely, a traditional classifier might process articles quickly but miss subtle contextual cues that indicate economic direction. This exemplifies the broader challenge we address: how can we develop classifiers that capture the nuanced understanding of LLMs while maintaining the efficiency necessary for real-time, large-scale deployment?

To address this problem, two primary categories of models have been widely adopted: large-scale transformer models and traditional machine learning algorithms. Transformer architectures, first introduced in \cite{vaswani2017attention} and popularized by Large Language Models (LLMs) like GPT, Claude, and Gemini, represent the state-of-the-art in performance \cite{wu2025advancementsnaturallanguageprocessing}. By leveraging complex self-attention mechanisms and deep semantic embeddings, they achieve a nuanced understanding of language that often translates to superior classification accuracy. However, this power comes at a steep price. Their immense size, with billions of parameters, makes both training and inference exceedingly slow and expensive, hindering their widespread adoption for many practical applications. In contrast, traditional machine learning algorithms such as Support Vector Machines (SVMs) \cite{cortes1995supportvector}, Gradient-Boosting Decision Trees (GBDTs) \cite{friedman2001greedy}, or Random Forests \cite{breiman2001randomforests} are significantly more efficient, offering rapid training and classification at a fraction of the cost. More importantly, their decisions are far more interpretable, a critical feature in domains where justifying a model's reasoning is paramount. Yet, these models typically 
%rely on statistical tokenization rather than semantic embeddings, 
requires domain specific supervision and has much smaller and simpler model structure,
which can limit their ability to capture the complex relationships within text, often leading to lower accuracy compared to LLMs.

A promising approach to bridge this gap is Knowledge Distillation (KD), a technique where a large, high-performing ``teacher'' model (the LLM) is used to train a smaller, more efficient ``student'' model (the traditional ML algorithm) \cite{hinton2015distillingknowledgeneuralnetwork, ghorbani2020surveykd, tang2019distillingbert}. The goal is to transfer the teacher's sophisticated ``knowledge'' to the student, thereby combining the high accuracy of an LLM with the efficiency and interpretability of a classical algorithm. However, a major bottleneck persists: the distillation process itself. Typically, it requires the expensive teacher model to label a massive dataset to create the training curriculum for the student. This step consumes significant computational resources and incurs high financial costs from API calls, undermining the very efficiency that KD aims to achieve.

Fortunately, this challenge of minimizing labeling costs by selecting only the most valuable data points is precisely the problem addressed by the field of active learning (AL) \cite{ren2021surveydeepactive}. The core idea of active learning is to allow a machine learning algorithm to intelligently choose the data from which it learns \cite{tong2001svmactive}. Rather than passively receiving a large, randomly selected training set, an active learning system iteratively queries an oracle (in our case, the teacher LLM) to label only the most informative unlabeled samples. By focusing the labeling effort on instances the model is most needed, AL has the potential to achieve high accuracy with a fraction of the labeled data required by traditional methods \cite{eindor2020activelearningbert}.

In this paper, we propose a novel approach that combines the principles of Knowledge Distillation with an intelligent active learning strategy called M-RARU (Multi-class Randomized Accept/Reject Uncertainty Sampling). Our approach works within an iterative loop: the student model first identifies a pool of candidate samples it is most uncertain about. Then, M-RARU's accept-reject mechanism strategically selects a subset of these candidates to be sent to the LLM teacher for labeling. This ensures that only the most valuable examples are used for training, dramatically improving the efficiency of the knowledge transfer process. This approach ensures the final student model is not only accurate but also retains the speed, cost-effectiveness, and interpretability of traditional machine learning.

We experimentally evaluated M-RARU against a standard random sampling baseline on multiple benchmark datasets, using five different student models (Support Vector Machine (SVM), Linear Discriminant Analysis (LDA), Random Forest (RF), Gradient Boosted Decision Tree (GBDT), and DistilBERT \cite{sanh2019distilbert}). The experimental results show that student models trained with M-RARU substantially outperform their randomly-sampled counterparts in accuracy and balanced accuracy. More importantly, M-RARU achieves this superior performance while drastically reducing the number of required teacher labels, leading to substantial savings in financial costs and overall training time. The resulting student models also offer much faster inference, providing a practical path to harness LLM power in resource-constrained applications.

Specifically, our contributions in this paper are as follows:
\begin{itemize}
\item We propose a novel approach that hybridizes Knowledge Distillation with Active Learning to address the high cost of training performant classifiers, efficiently leveraging an LLM teacher to train a smaller student model \cite{dasgupta2022distilling}.
\item We introduce Multi-class Randomized Accept/Reject Uncertainty Sampling (M-RARU), a specific AL algorithm that intelligently selects data to create a small yet highly effective training set, maximizing student model performance while minimizing LLM labeling costs.
\item We conduct extensive experiments on multiple large, real-world text corpora, demonstrating that our proposed method substantially outperforms a random sampling baseline across a diverse set of student models, verifying our approach as a practical path to developing fast, accurate, and cost-effective classifiers.
\end{itemize}

The rest of the paper is structured as follows. Section \ref{sec:bg} introduces the background and problem definitions. Section \ref{sec:approach} presents our solutions. Section \ref{sec:exp} describes the experimental environment and presents the evaluation results. Section \ref{sec:rw} describes works that are closely related to us. Finally, Section \ref{sec:con} concludes.
\section{Problem Definition \& Background}
\label{sec:bg}

In this section, we formally introduce our problem and provide the necessary background for our approach.

\subsection{Knowledge Distillation Task}
To frame the knowledge distillation task addressed in this work, we consider a scenario involving high-dimensional text data. Each data item (e.g., a sentence or document) is represented as a high-dimensional vector via an embedding model. A large, complex ``teacher" model, which has high performance but is computationally expensive, already exists. The primary challenge is to train a smaller, more efficient ``student" model to replicate the teacher's predictive capabilities. Consequently, the goal of our active learning approach is to strategically select a small, highly informative subset of unlabeled data for the teacher to label. This dataset is then used to train the student model, aiming to achieve performance comparable to the teacher with minimal labeling cost.

\subsection{Problem Settings} %NEED TO EDIT; I tried my best
\label{sec:problem_settings}
To formalize the knowledge distillation problem addressed in this work:
\\
Consider a $d$-dimensional data space $D$ containing $N$ data items, where each item belongs to one of $C$ possible classes. This formulation targets both binary ($C=2$) and multi-class ($C>2$) classification tasks.
\\
Further, consider a powerful ``teacher" model, $M_T$, which can provide a high-quality class label for any item in $D$, and a smaller ``student" model, $M_S$, that we aim to train.
\\
The training process uses a small subset of data, $L \subset D$, of size $n$ (where $n \ll N$), which is interactively selected from the unlabeled pool and labeled by the teacher model $M_T$.
\\
The objective is to construct a student model $M_S$ that accurately predicts the class labels for the entire dataset $D$, effectively mimicking the behavior of $M_T$, by using a query strategy to build the most informative training set $L$.
\\
The success of this knowledge transfer is measured by the predictive performance of the student model. We focus on \emph{accuracy} and \emph{balanced accuracy} as they are particularly well-suited for this task.
\begin{itemize}
    \item \textbf{Accuracy} is the most direct measure of performance, defined as the proportion of all data items that are correctly classified. It provides a clear, overall assessment of the model's correctness.
    \item \textbf{Balanced Accuracy} is crucial in scenarios with imbalanced class distributions, which are common in real-world text datasets. It is calculated as the average of the recall for each class, ensuring that the student model is evaluated fairly across all classes and not rewarded for simply predicting the majority class.
\end{itemize}
Our goal is to design a data selection solution that maximizes these measures for a fixed budget of $n$ labels provided by the teacher.

\begin{comment}
Given the fact that the user is unaware of the proper specification that describes $\Re^+$, the user can only recognize it in hindsight based on the data items predicted as relevant and retrieved by $\rho$. Hence, the objective is to predict accurately all regions in $\Re^+$, which can be naturally measured using the \emph{$F$-measurement} \cite{AIDE,request}, the harmonic mean between precision and recall. Particularly, for q data space $D$ of size $N$ and a predictive model $\rho$, all the data items in $D$ predicted by $\rho$ as positive should be relevant to user exploration, and the remaining data items in $D$ should be irrelevant.  Accordingly, our goal is to design a solution that would maximize the $F$-measure for a fixed amount of user labels, such that $F$-measure is defined as: 

\begin{equation}
F(N)=\frac{2\cdot\text{Precision}(N)\cdot\text{Recall}(N)}{\text{Precision}(N)+\text{Recall}(N)}
\end{equation}

Here, precision measures the portion of true relevant data items among all the data items predicted as relevant by $\rho$.  
%
Hence, true relevant, or true positive, indicates that a data item is both relevant to the user and has been predicted as relevant by $\rho$.  
%
If a data item is irrelevant but predicted to be relevant by $\rho$, it is considered a false positive.   
% 
Recall measures the ratio of the true relevant data items captured by $\rho$ to all the data items that are actually relevant to the user.  
\end{comment}

\subsection{Active Learning}
Active learning is a paradigm in machine learning that aims to achieve high accuracy while minimizing the amount of labeled data required for training (\cite{FAN2024122356}). It employs query strategies to iteratively select the most informative unlabeled sample (i.e., data object) from unlabeled data, obtain the true labels from an expert source (in our case an LLM), and then update the model with this new information. The query strategy dictates how data points/informational inputs are chosen.

Numerous query strategies \cite{Settles10activelearning} have been proposed to define the ``informativeness" of samples in the literature,
including: {\em Uncertainty Sampling}, {\em Query-By-Committee}, {\em Expected Model Change}, {\em Expected Error Reduction}, and {\em Expected Model Output Change}.
Among these query strategies, Uncertainty Sampling is the most commonly used one because of its simplicity and efficiency, as pointed out in \cite{Settles10activelearning}.

\subsubsection*{Uncertainty Sampling}
Uncertainty sampling \cite{Lewis94a} is a query strategy that can be used with any probability-based classification model (Naive Bayes, SVM, etc..). It selects samples based on the model's uncertainty about their classification (\cite{electronics11030396}).
The intuition underlying uncertainty sampling is that patterns with high uncertainty are hard to classify,
so obtaining high-uncertainty labels boosts accuracy of classification models (more than say, random sampling).

Particularly, in classification models (e.g., with class labels a, b, c, and d), the most uncertain example $\textbf{x}$ is the one which can be assigned to any class label $z(\textbf{x})$ with an even probability distribution (e.g., 0.25, 0.25, 0.25, 0.25).

Inspired by the idea of uncertainty, also known as {\em least confidence}, \cite{Lewis94a} proposes a measurement of uncertainty for binary classification models, which easily extends to categorical classification models:
\begin{equation}
\label{uncertaintymeasure}
u^{(lc)}(\textbf{x})=1-p(\hat{y}|\textbf{x})
\end{equation}
where $u^{(lc)}(\textbf{x})$ is the uncertainty score with the least confidence measurement of \textbf{x}, and $\hat{y}$
is
%%means
the predicted class label of the unlabeled $\textbf{x}$.
Accordingly, after measuring the uncertainty of each unlabeled sample, the unlabeled sample with highest uncertainty is selected:
\begin{equation}
\textbf{x}^*=\text{argmax}_{\textbf{x}}u(\textbf{x})
\end{equation}
where $u(\textbf{x})$ can be any other measurement of informativeness over the unlabeled sample \textbf{x}.

\section{Our Approach}
\label{sec:approach}

In this section, we formally describe our proposed framework, which integrates active learning with knowledge distillation to produce efficient and accurate classifiers.

\begin{algorithm}[t]
\caption{The Knowledge Distillation Process}
\label{alg:distillation}
{\small
\begin{algorithmic}[1]
\REQUIRE The raw text corpus $D$, a teacher model $M_T$
\ENSURE A trained student model $M_S$
\STATE Convert $D$ into a set of embeddings $E$
\STATE $L \leftarrow \emptyset$ \COMMENT{Initialize the training set for the student}
\STATE $U \leftarrow E$ \COMMENT{Initialize the unlabeled pool}
\STATE $M_S \leftarrow$ initialize student model
\WHILE{$U$ is not empty}
\STATE Randomly select one sample $x$ from $U$
\STATE Solicit normalized uncertainty score $p$ for $x$ from $M_S$
\STATE With probability $p$, add $x$ to the labeling set $L$
\STATE $U \leftarrow U - \{x\}$
\ENDWHILE
\STATE Request labels for all samples in $L$ from teacher model $M_T$
\STATE Train student model $M_S$ on the labeled set $L$
\STATE Return trained student model $M_S$
\end{algorithmic}
}
\vspace{-5pt}
\end{algorithm}

\subsection{Proposed Solution}

Our proposed solution is designed to bridge the gap between the high performance of Large Language Models (LLMs) and the efficiency of traditional machine learning classifiers. The framework aims to achieve two primary goals: 1) minimize the financial and computational cost associated with using an LLM ``teacher'' for labeling, and 2) train a smaller ``student'' model that achieves the highest possible accuracy by learning from a strategically selected, information-rich dataset.

As illustrated in Algorithm \ref{alg:distillation}, our framework identifies the most valuable data for training through an iterative selection process. The system first converts the entire raw text corpus into a set of numerical vector representations, or embeddings, to make the data processable by machine learning models (Line 1). It then initializes an empty training set $L$ and a student model $M_S$ (Lines 2-4). The core of our approach is a loop that intelligently builds the training set $L$ (Lines 5-10). In each iteration, instead of exhaustively searching the entire unlabeled pool, the framework randomly selects a data sample and queries the current student model for its predictive uncertainty. This uncertainty score is then used to probabilistically decide whether the sample is informative enough to be added to the set $L$ for later labeling by the teacher. This process continues until every sample in the original corpus has been considered.

Once the selection phase is complete, the framework sends only the curated, high-value samples in set $L$ to the powerful but expensive LLM teacher to obtain high-quality labels (Line 11). This small, targeted training set is then used to train the final student model (Line 12). By focusing the teacher's effort exclusively on the most informative examples, our framework facilitates an efficient knowledge transfer, producing a student model that emulates the teacher's performance at a fraction of the cost.

A key advantage of this approach is the enhanced interpretability of the final student model. While LLMs and even DistilBERT operate as complex "black boxes," the decision-making processes of models like GBDT, Random Forest, and SVM can be readily explained using well-established techniques such as SHAP (SHapley Additive exPlanations) \cite{lundberg2017unified} or LIME (Local Interpretable Model-agnostic Explanations) \cite{ribeiro2016should}. These methods can generate feature-level explanations for individual predictions, revealing which words or phrases most influenced a particular classification. This transparency is invaluable in high-stakes domains like finance or regulation, where understanding \textit{why} a model made a certain decision is as important as the decision itself. 

In the following sections, we will present each main component of our approach in detail.

\subsection{Data Embedding}

In the domain of natural language processing, the representation of text data is a critical first step that profoundly influences the performance of any machine learning model. To this end, embedding methods are employed to transform unstructured text into dense numerical vectors that capture semantic relationships. These methods aim to create feature representations such that the proximity between vectors in the learned vector space reflects the semantic similarity of the corresponding text in its original form.

A large variety of algorithms have been proposed for this task. Well-recognized approaches such as Word2vec \cite{DBLP:conf/nips/MikolovSCCD13}, GloVe \cite{DBLP:conf/emnlp/PenningtonSM14}, and FastText generate embeddings at the word level, while more advanced transformer-based models like BERT or sentence encoders like the Universal Sentence Encoder \cite{DBLP:conf/acl/YangCAGLCAYTSSK20} create contextualized representations for entire sentences or documents. These methods provide rich representations that preserve the nuances of linguistic context, enabling classifiers to perform complex reasoning.

In our work, for student models that requires embeddings, we leverage sentence-level embeddings to ensure that the full semantic meaning of each text sample is captured. Using a single, unified embedding method for all traditional student models also ensures consistency and comparability of results, as different embedding techniques can produce vectors of varying dimensionality (from hundreds to thousands of dimensions), which could otherwise introduce confounding variables into our performance evaluation.

\subsection{Query Strategy}

The \emph{Query Strategy} is the component of our framework responsible for minimizing the labeling cost while maximizing the student model's ultimate accuracy. In the context of our approach, a ``query'' refers to the process of selecting an unlabeled data sample to be labeled by the teacher LLM. Our framework leverages a specialized form of uncertainty sampling to intelligently build the training set and guide the knowledge distillation process.

\subsubsection*{Uncertainty Sampling}

Uncertainty sampling is a widely adopted active learning strategy predicated on a simple yet powerful intuition: a model gains the most information from samples it is least certain about. By prioritizing these ambiguous samples for labeling, a model can resolve confusion at its decision boundary more quickly, leading to faster convergence and higher accuracy with fewer labeled examples. To measure the uncertainty of a data object $x$, a probabilistic predictive model is needed to report the probability of $x$ belonging to each possible class. The sample for which the model's prediction is least confident (e.g., closest to a 50\% probability in a binary task) is considered the most uncertain and, therefore, the most informative.

\subsubsection*{Challenges with Traditional Uncertainty Sampling}

Despite its effectiveness, traditional uncertainty sampling suffers from two major drawbacks, particularly in the context of large datasets: 1) \textbf{shortsightedness} \cite{albook} and 2) \textbf{low scalability} \cite{request}.

\textit{Shortsightedness} arises because the model's uncertainty is estimated using only the information from the few samples it has already seen. This can create a bias, causing the strategy to repeatedly select samples clustered around a single, noisy region of the decision boundary while ignoring other potentially informative areas of the feature space. \textit{Low scalability} is a computational bottleneck; conventional uncertainty sampling requires an \textbf{exhaustive search} over the entire unlabeled dataset in every iteration to find the single most uncertain sample. This process incurs prohibitive processing costs and introduces significant delays, making it impractical for large-scale applications.

\subsubsection*{Randomized Uncertainty} 
To overcome the first drawback mentioned above, the work in \cite{Yanbing} combines uncertainty with some degree of randomness. 
In particular, an unlabeled object that would be presented to the user as an example is probabilistically selected from the entire set of unlabeled objects. 
This probabilistic framework requires that the 'informativeness' of each sample be a non-negative, quantitatively meaningful score suitable for normalization. Because uncertainty scores are derived directly from model probabilities, they are a natural fit for creating such a selection distribution, a property not guaranteed by all informativeness metrics used in active learning \cite{Settles10activelearning}.
The probability that an unlabeled object $x$ is selected is proportional to its uncertainty score: 
\begin{equation}
p(x\mbox{ is selected})=\frac{u(x)}{\sum_{x_u\in U}u(x_u)}
\end{equation}
where $U$ is the set of unlabeled objects and $u(\textbf{x})$ is the uncertainty score of \textbf{x}. 

Since the probability that an unlabeled object $x$ is chosen as an example is equal to its normalized uncertainty score, 
therefore, less uncertain objects can still have a small chance of being accepted as examples, which essentially reduces the bias introduced by the labeled samples.

\subsubsection*{Multi-class Randomized Accept/Reject Uncertainty Sampling (M-RARU)}
While the Randomized Uncertainty strategy addresses traditional uncertainty sampling's drawback of shortsightedness, the issue of low scalability still remains. To overcome this limitation, the work in \cite{request} and \cite{exnav} introduced a randomized Accept/Reject mechanism that allows uncertainty estimation to be performed efficiently for binary classifications. 
However, many real-world classification tasks often involve multiple classes or labels. Therefore, methods designed only for binary classification are not suitable for these knowledge distillation tasks. In this work we introduce the Multi-class Randomized Accept/Reject Uncertainty Sampling (M-RARU). M-RARU addresses both shortsightedness and scalability for both binary and multi-class classification tasks by introducing a randomized, probabilistic selection mechanism eliminates the need to perform exhaustive search over the entire data space. Particularly, in each step, M-RARU randomly selects a single sample from the unlabeled pool, calculates its uncertainty score, and then uses this score to make a probabilistic decision on whether to “accept” the sample for labeling or “reject” it and move on.

The probability of an unlabeled data sample $\mathbf{x}$ being accepted into the training set $L$ under M-RARU is defined as:
\begin{equation}
\label{raruaccept}
p(\mathbf{x} \text{ is accepted}) = 1 - \max_{k \in \{1,\ldots,K\}} \Pr(C_k|\mathbf{x})
\end{equation}
where $\Pr(C_k|\mathbf{x})$ is the probability of $\mathbf{x}$ being assigned the class label $C_k$ by the student model, and $K$ is the total number of classes. This formula directly captures the model's uncertainty: when the maximum predicted probability is low (indicating the model is uncertain about all classes), the acceptance probability is high. Conversely, when the model is confident in its prediction (high maximum probability), the acceptance probability is low. This ensures that highly uncertain samples have a high probability of being accepted, while still allowing less uncertain samples a chance to be selected, which helps mitigate the shortsightedness bias.
This formula is designed around the model's prediction confidence because the accept/reject mechanism requires an uncertainty score that can function as a direct probability of acceptance. Using the maximum prediction probability allows for the creation of a score naturally bounded within the required [0,1] range. In contrast, other common metrics like Shannon entropy produce a score on a different scale (e.g., [0,log(K)]), making them less compatible in this probabilistic decision framework.
By randomly visiting unlabeled objects until one is accepted, M-RARU provides an early termination to the costly exhaustive search, directly solving the scalability problem. This combination of randomization and uncertainty-based acceptance allows the framework to efficiently build a diverse and highly informative training set, preserving the core benefits of uncertainty sampling while adapting it for large-scale knowledge distillation.

\section{Experimental Evaluation}
\label{sec:exp}

In this section, we present the results of our experiments. We begin by introducing the experimental setup and then demonstrate the performance of our proposed scheme against the baseline across various student models and datasets.

\begin{table}[t]
\caption{EXPERIMENTAL PARAMETERS}
\centering
\small
  \begin{tabularx}{\linewidth}{ | l | X | }
    \hline
    \textbf{Parameter} & \textbf{Value} \\ \hline
    Experimental Datasets & Public Comments, LSEG Data \& Analytics. Global News Archive Database (GNAD) \\ \hline
    Data Objects (Public Comments) & 125,179 \\ \hline
    Data Objects (GNAD) & 12,288 \\ \hline
    Embedding Dimensions & 384 \\ \hline
    Embedding Model & all-MiniLM-L6-v2 \\ \hline
    Teacher Model (Oracle) & gemma-3-27b-it-qat-q4\_0-gguf \\ \hline
    Initial Labeled Pool & Randomly sampled until at least one sample per class is present \\ \hline
    AL Batch Size & 25 \\ \hline
    Max Labeled Examples & 6,275 (Public Comments), 6,150 (GNAD) \\ \hline
    Considered AL Schemes & M-RARU, Random Sampling (RANDOM) \\ \hline
    Student Models & SVM, LDA, RF, GBDT, DistilBERT \\ \hline
    Performance Measures & Accuracy, Balanced Accuracy \\ \hline
    Number of Runs per Result & 5 (1 for DistilBERT) \\
    \hline
  \end{tabularx}
  \label{t:params}
\end{table}

\subsection{Experiment Setup}

%===============================================
% PUBLIC COMMENTS RESULTS - All Models
%===============================================

% --- Row 1: SVM Public Comments (Acc & Bal Acc), LDA Public Comments (Acc & Bal Acc) ---
\begin{figure*}[!htb]
    \centering
    \begin{minipage}{0.24\textwidth}
        \centering
        \includegraphics[width=\linewidth]{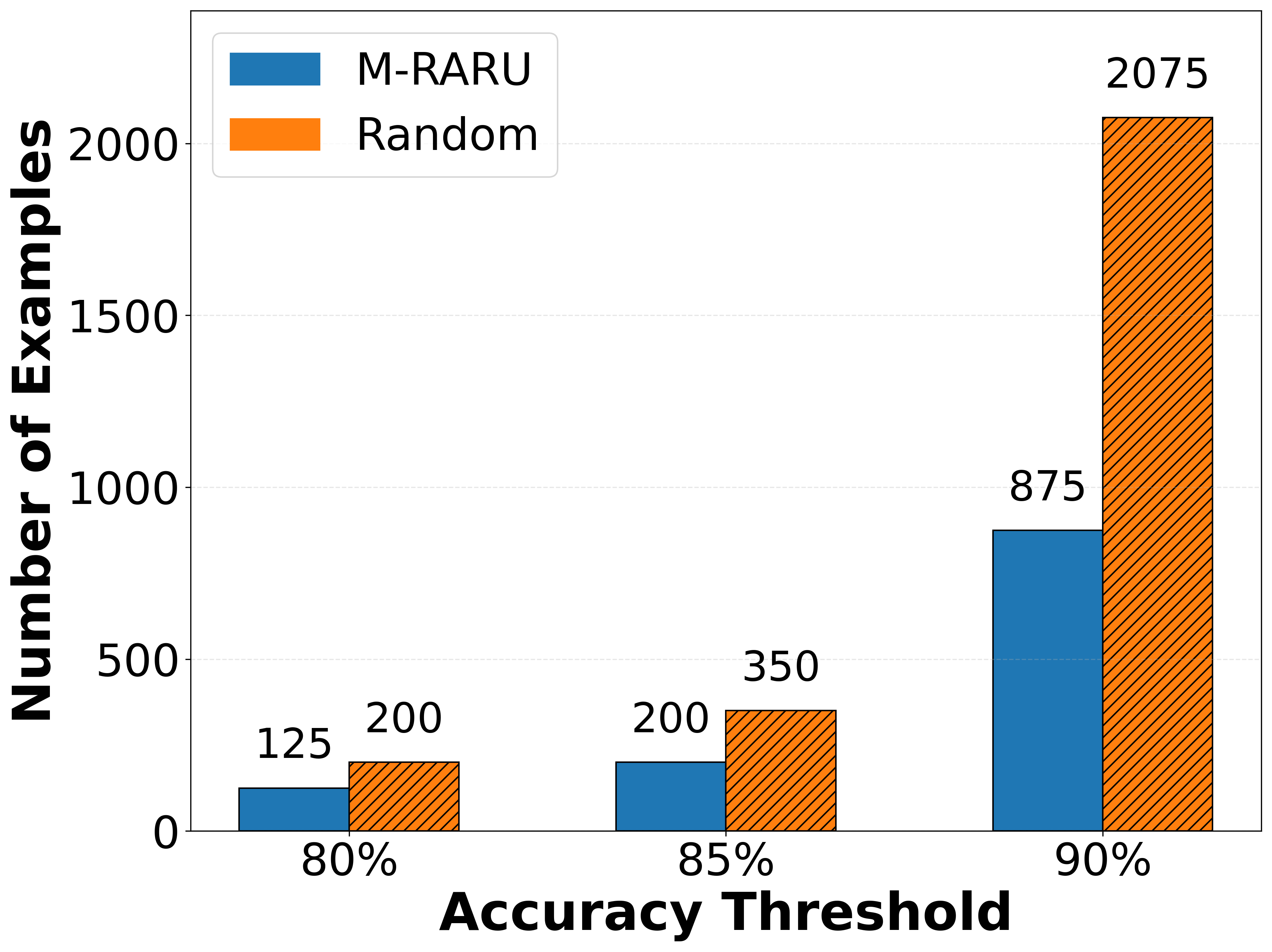}
        \caption{SVM Public Comments Accuracy}
        \label{fig:svmpubacc}
    \end{minipage}\hfill
    \begin{minipage}{0.24\textwidth}
        \centering
        \includegraphics[width=\linewidth]{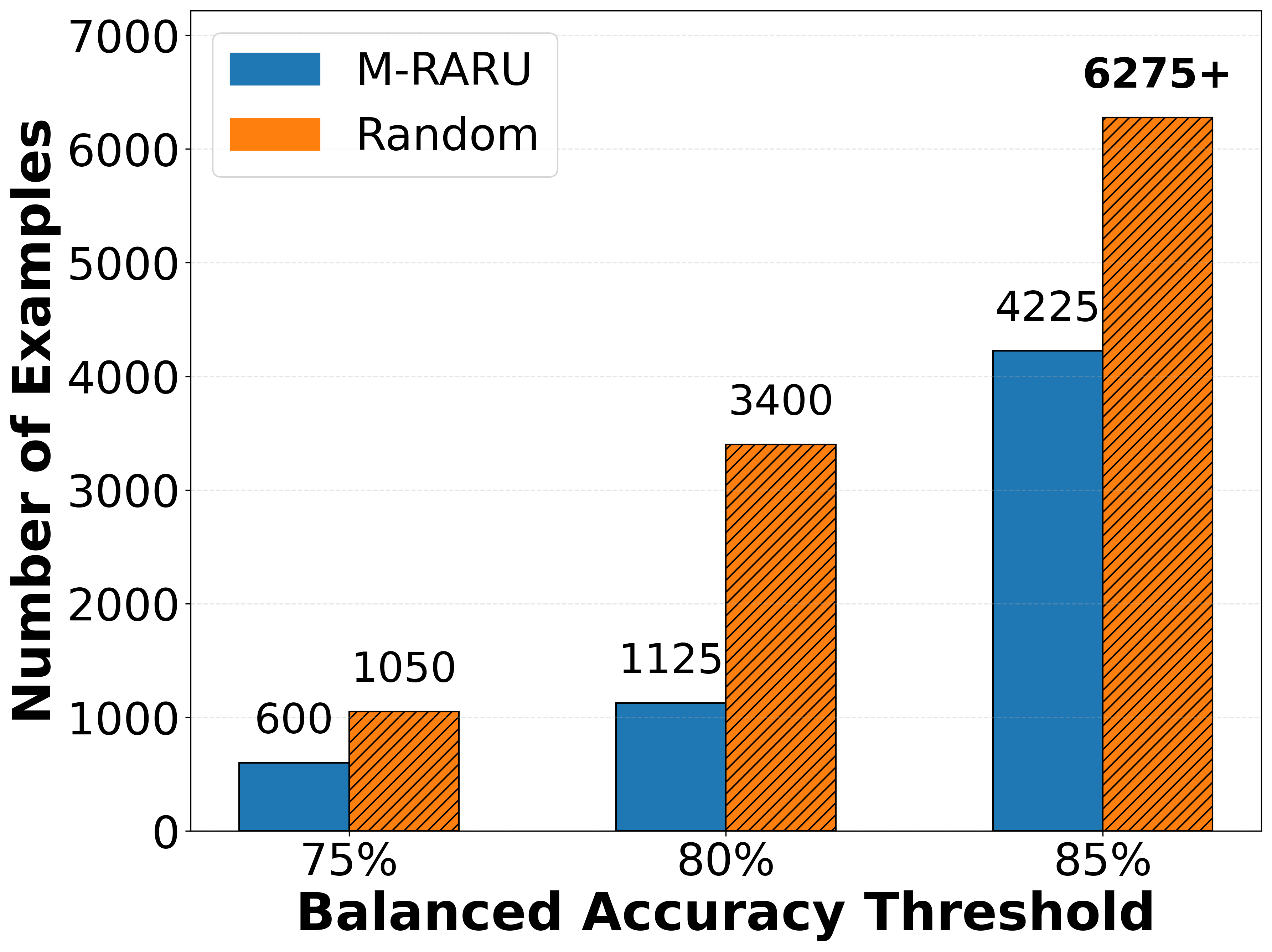}
        \caption{SVM Public Comments Balanced Accuracy}
        \label{fig:svmpubbal}
    \end{minipage}\hfill
    \begin{minipage}{0.24\textwidth}
        \centering
        \includegraphics[width=\linewidth]{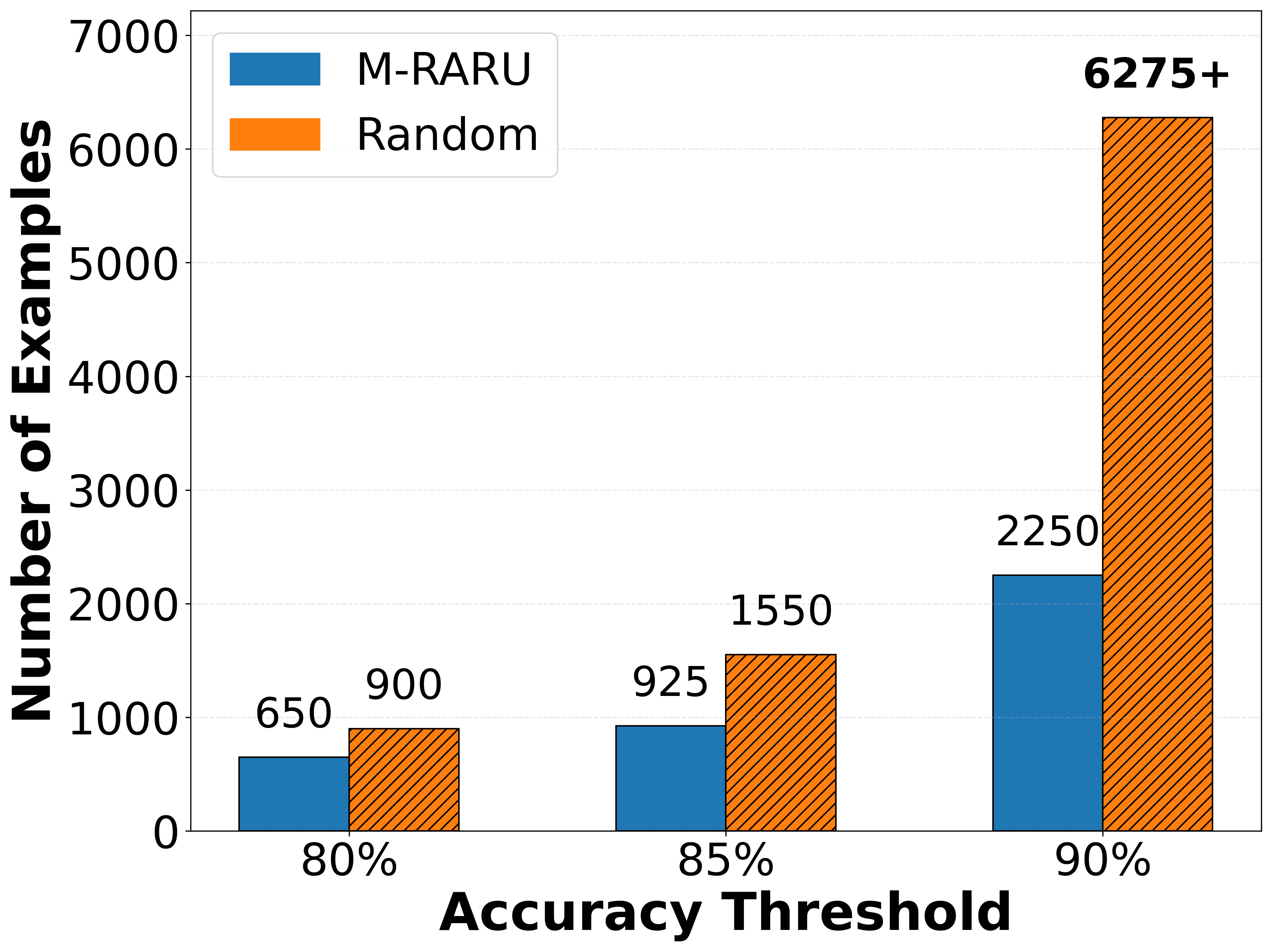}
        \caption{LDA Public Comments Accuracy}
        \label{fig:ldapubacc}
    \end{minipage}\hfill
    \begin{minipage}{0.24\textwidth}
        \centering
        \includegraphics[width=\linewidth]{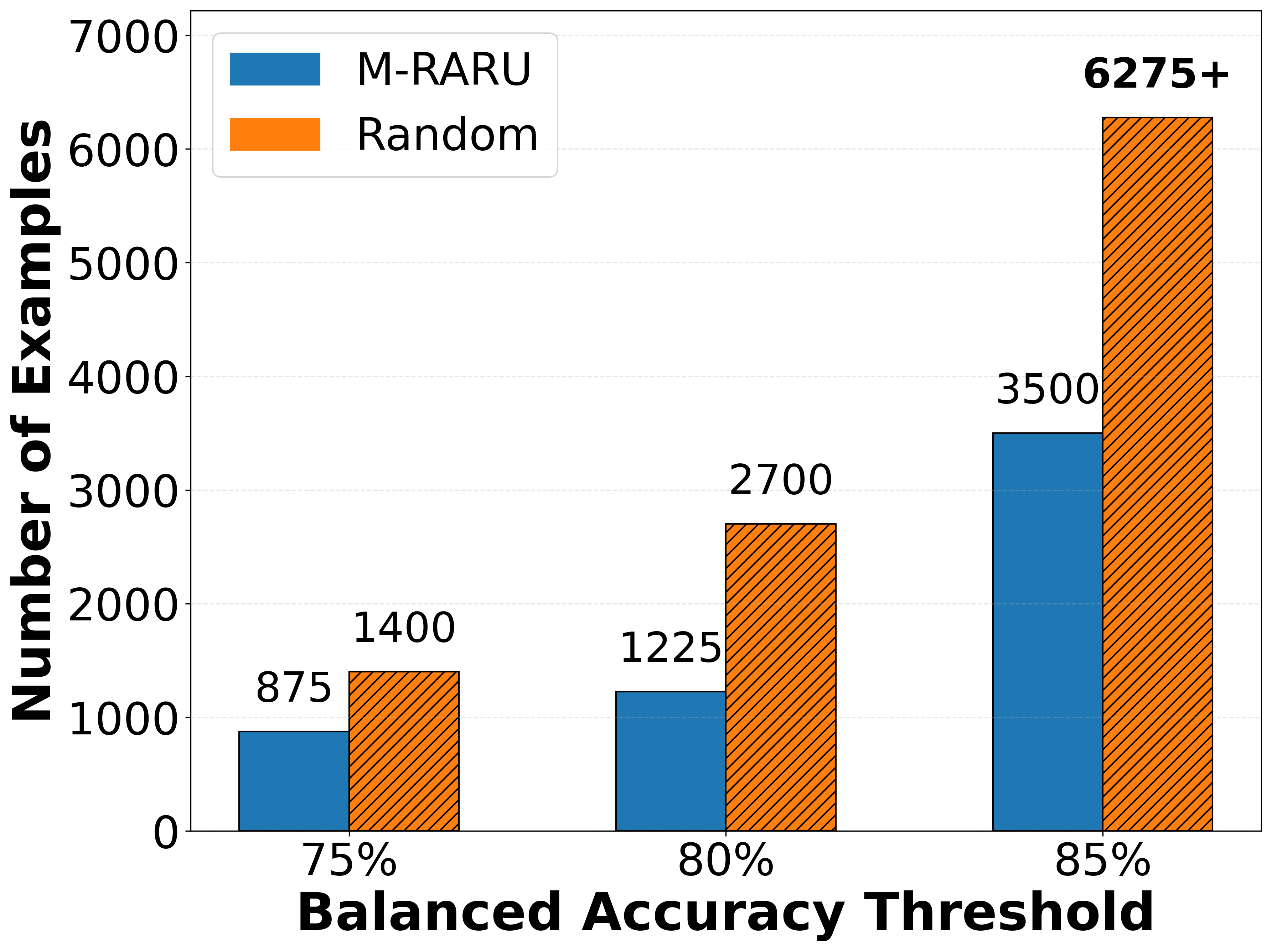}
        \caption{LDA Public Comments Balanced Accuracy}
        \label{fig:ldapubbal}
    \end{minipage}
\end{figure*}

% --- Row 2: RF Public Comments (Acc & Bal Acc), GBDT Public Comments (Acc & Bal Acc) ---
\begin{figure*}[!htb]
    \centering
    \begin{minipage}{0.24\textwidth}
        \centering
        \includegraphics[width=\linewidth]{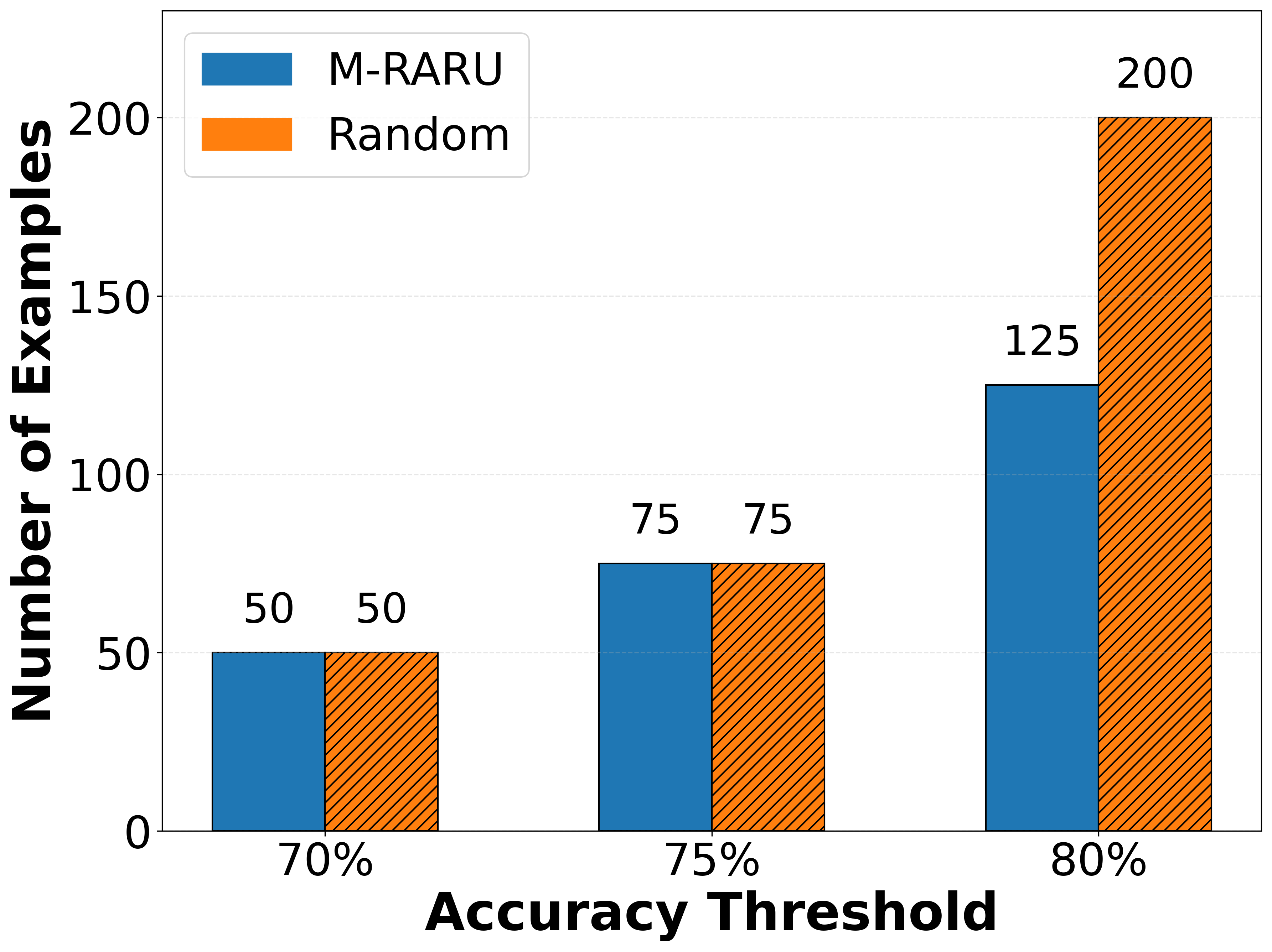}
        \caption{RF Public Comments Accuracy}
        \label{fig:rfpubacc}
    \end{minipage}\hfill
    \begin{minipage}{0.24\textwidth}
        \centering
        \includegraphics[width=\linewidth]{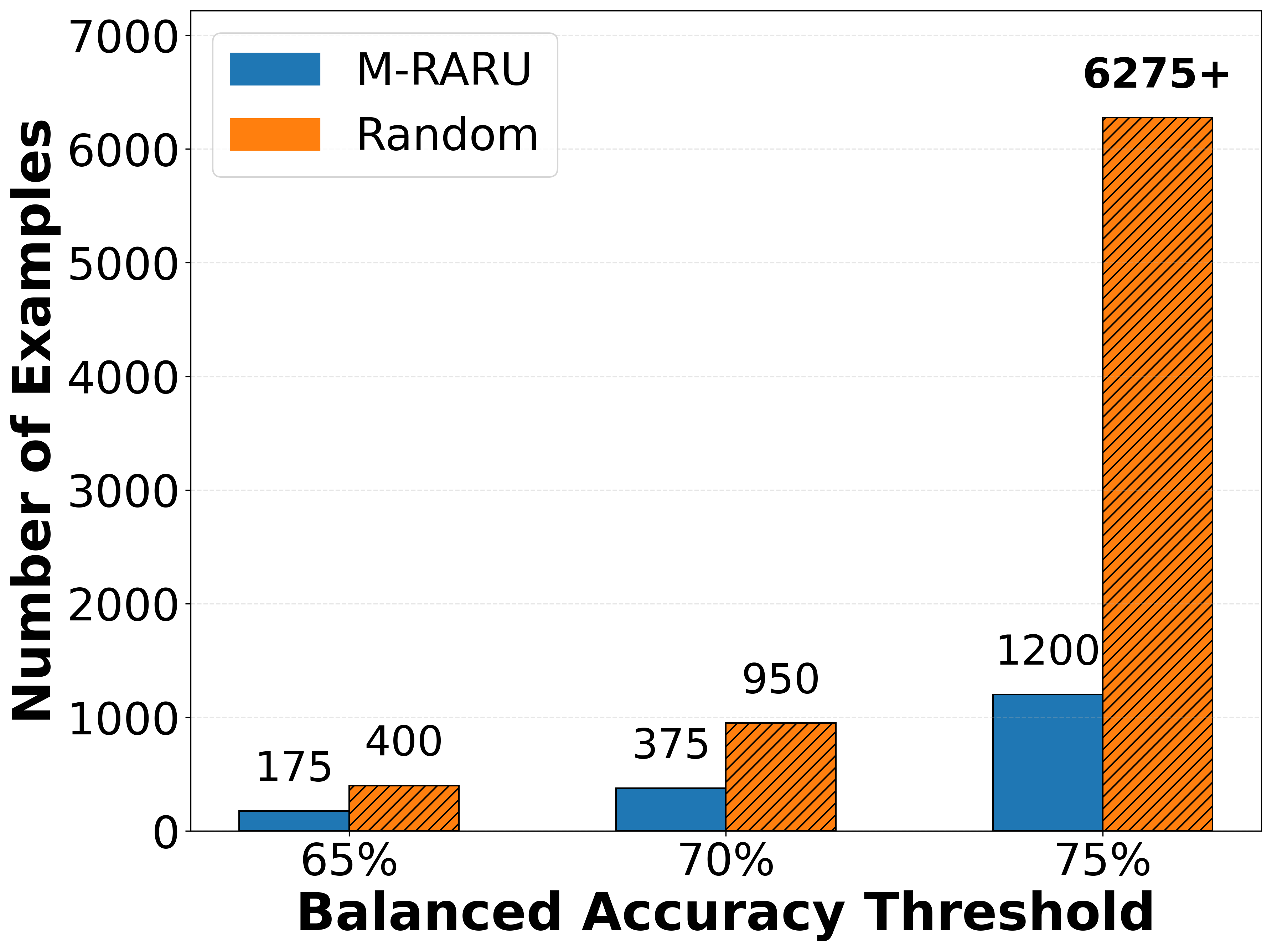}
        \caption{RF Public Comments Balanced Accuracy}
        \label{fig:rfpubbal}
    \end{minipage}\hfill
    \begin{minipage}{0.24\textwidth}
        \centering
        \includegraphics[width=\linewidth]{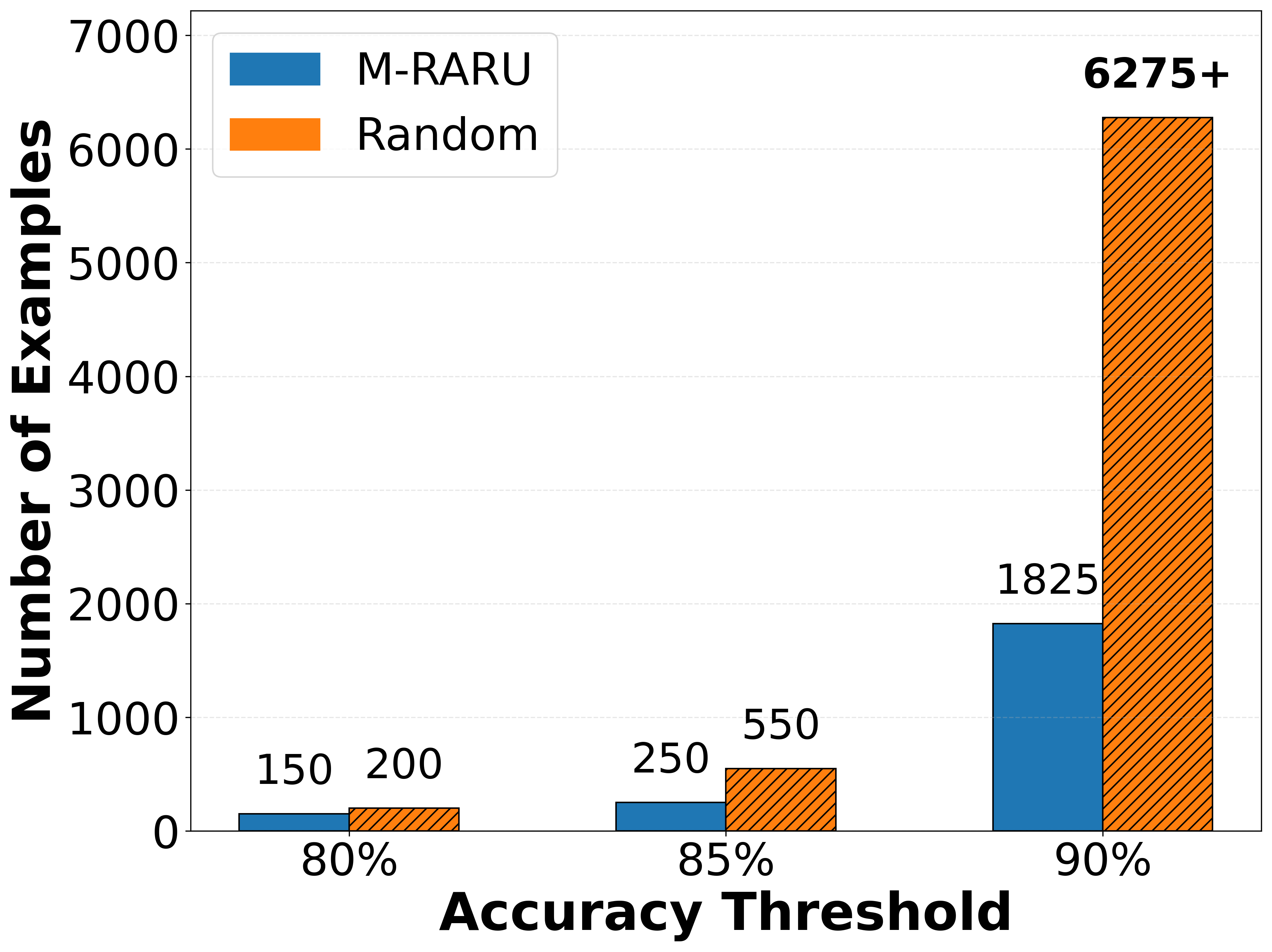}
        \caption{GBDT Public Comments Accuracy}
        \label{fig:gbdtpubacc}
    \end{minipage}\hfill
    \begin{minipage}{0.24\textwidth}
        \centering
        \includegraphics[width=\linewidth]{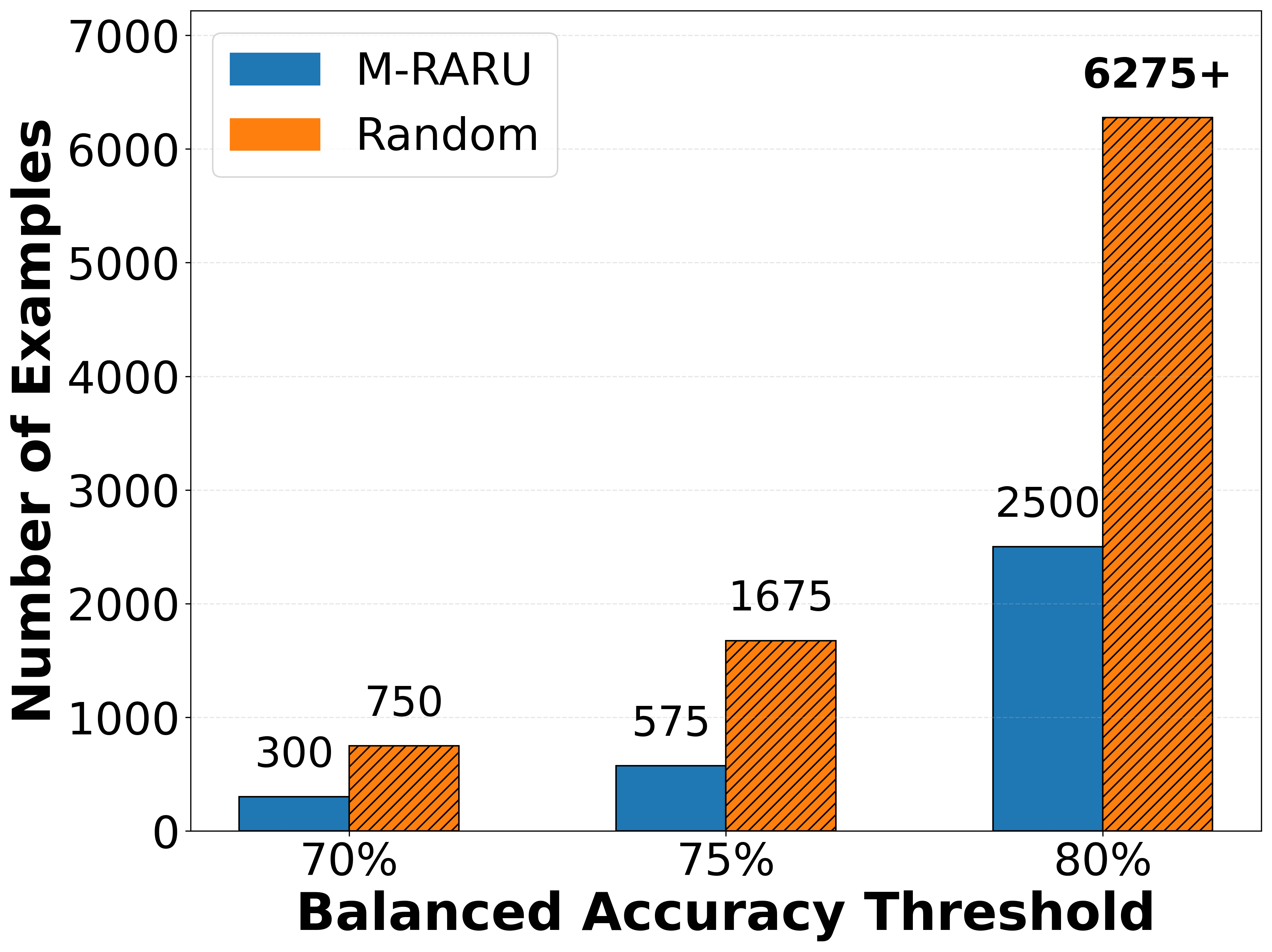}
        \caption{GBDT Public Comments Balanced Accuracy}
        \label{fig:gbdtpubbal}
    \end{minipage}
\end{figure*}

% --- Row 3: DistilBERT Public Comments (Acc & Bal Acc), plus two empty slots or start of GNAD ---
\begin{figure*}[!htb]
    \centering
    \begin{minipage}{0.24\textwidth}
        \centering
        \includegraphics[width=\linewidth]{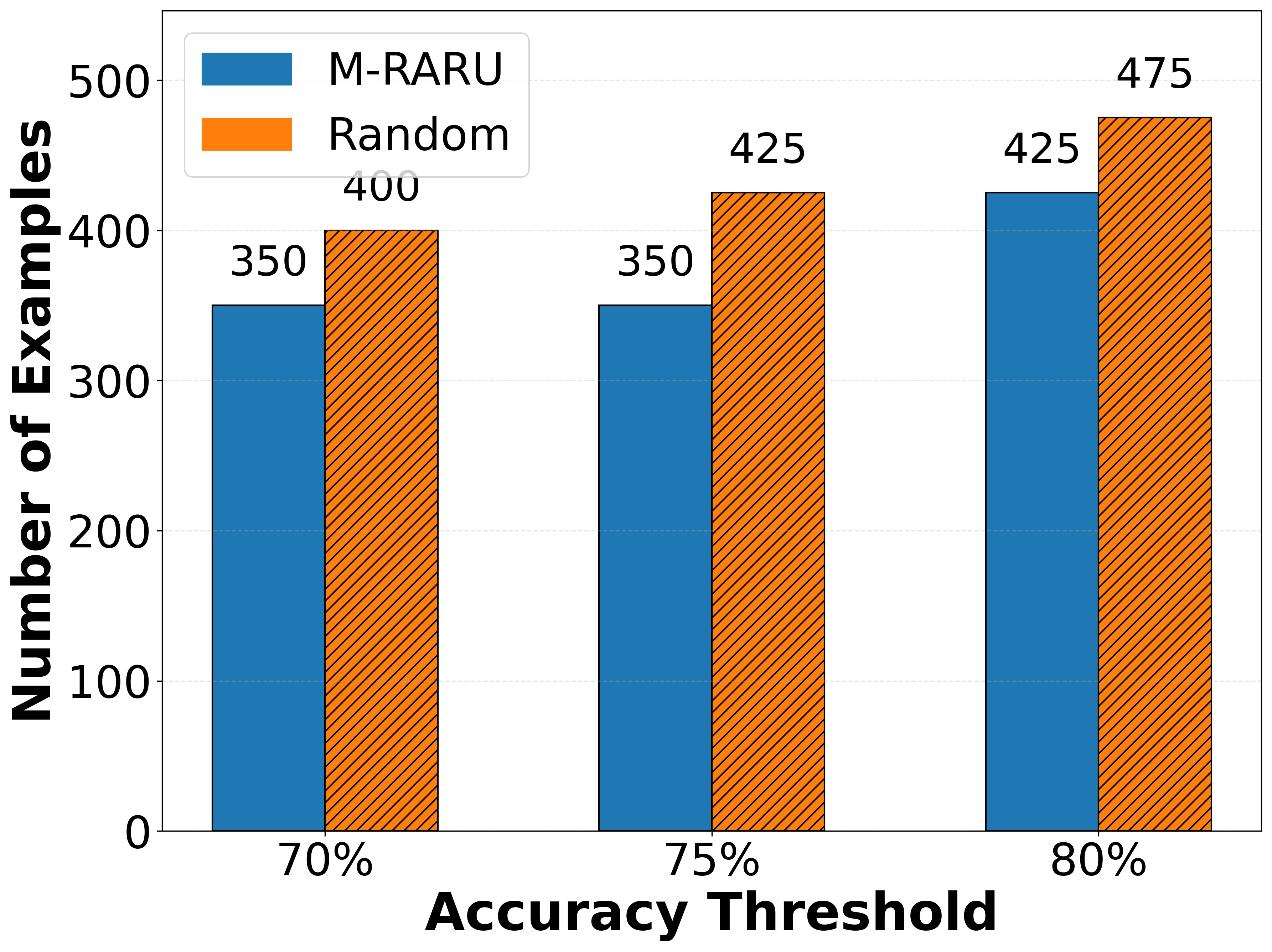}
        \caption{DistilBERT Public Comments Accuracy}
        \label{fig:bertpubacc}
    \end{minipage}\hfill
    \begin{minipage}{0.24\textwidth}
        \centering
        \includegraphics[width=\linewidth]{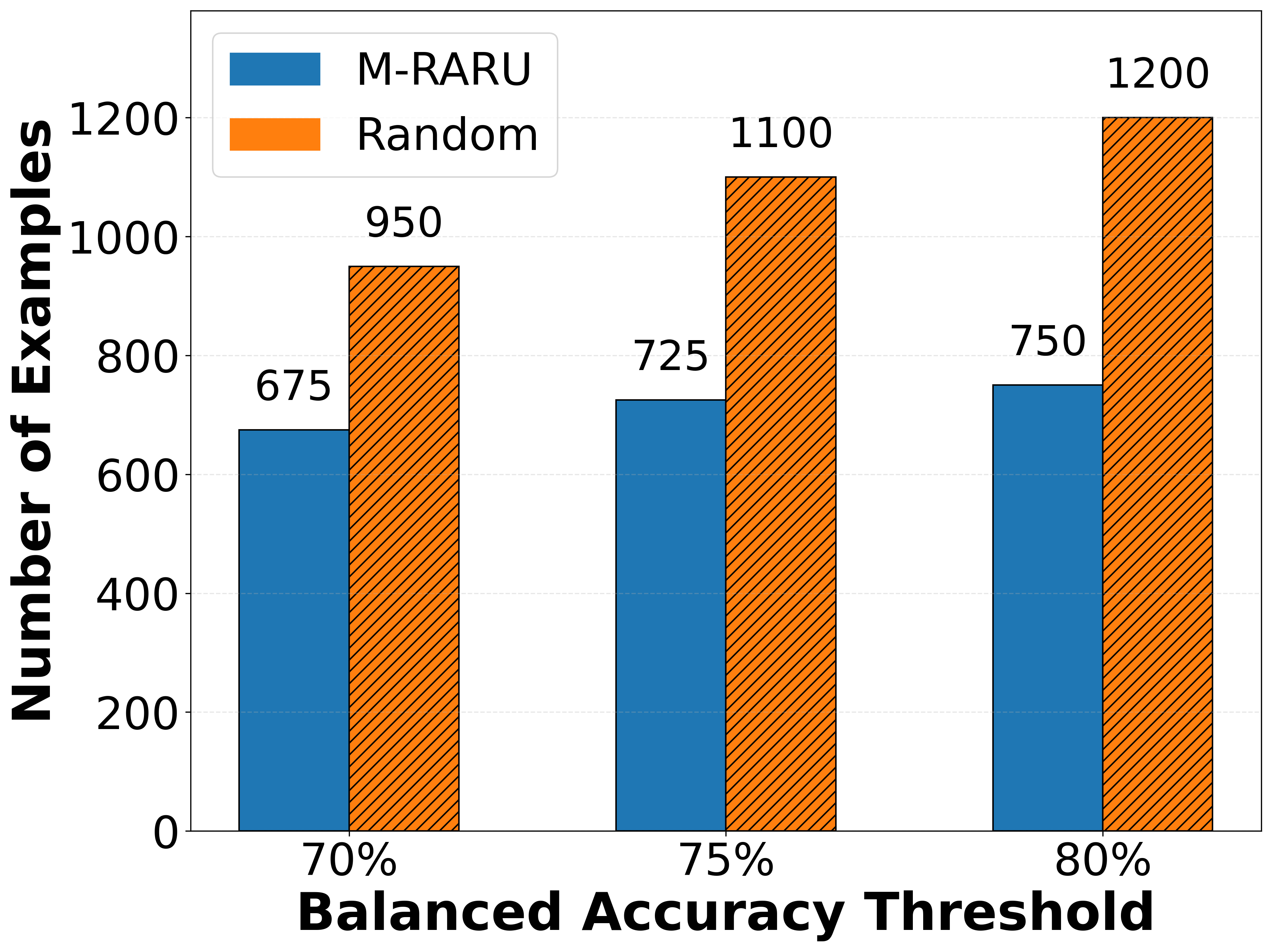}
        \caption{DistilBERT Public Comments Balanced Accuracy}
        \label{fig:bertpubbal}
    \end{minipage}\hfill
    \begin{minipage}{0.24\textwidth}
        \centering
        \includegraphics[width=\linewidth]{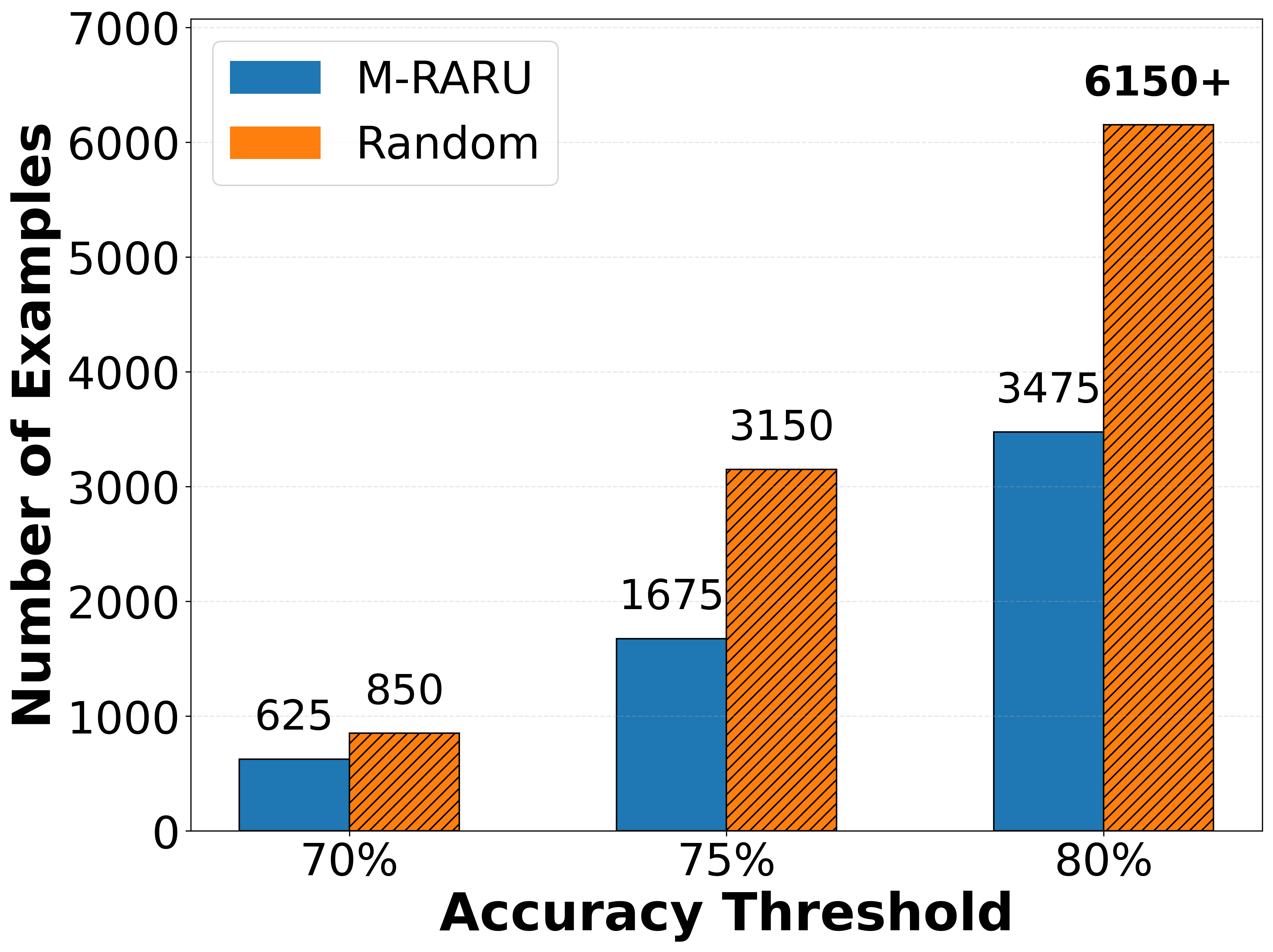}
        \caption{SVM GNAD Accuracy}
        \label{fig:svmtrnaacc}
    \end{minipage}\hfill
    \begin{minipage}{0.24\textwidth}
        \centering
        \includegraphics[width=\linewidth]{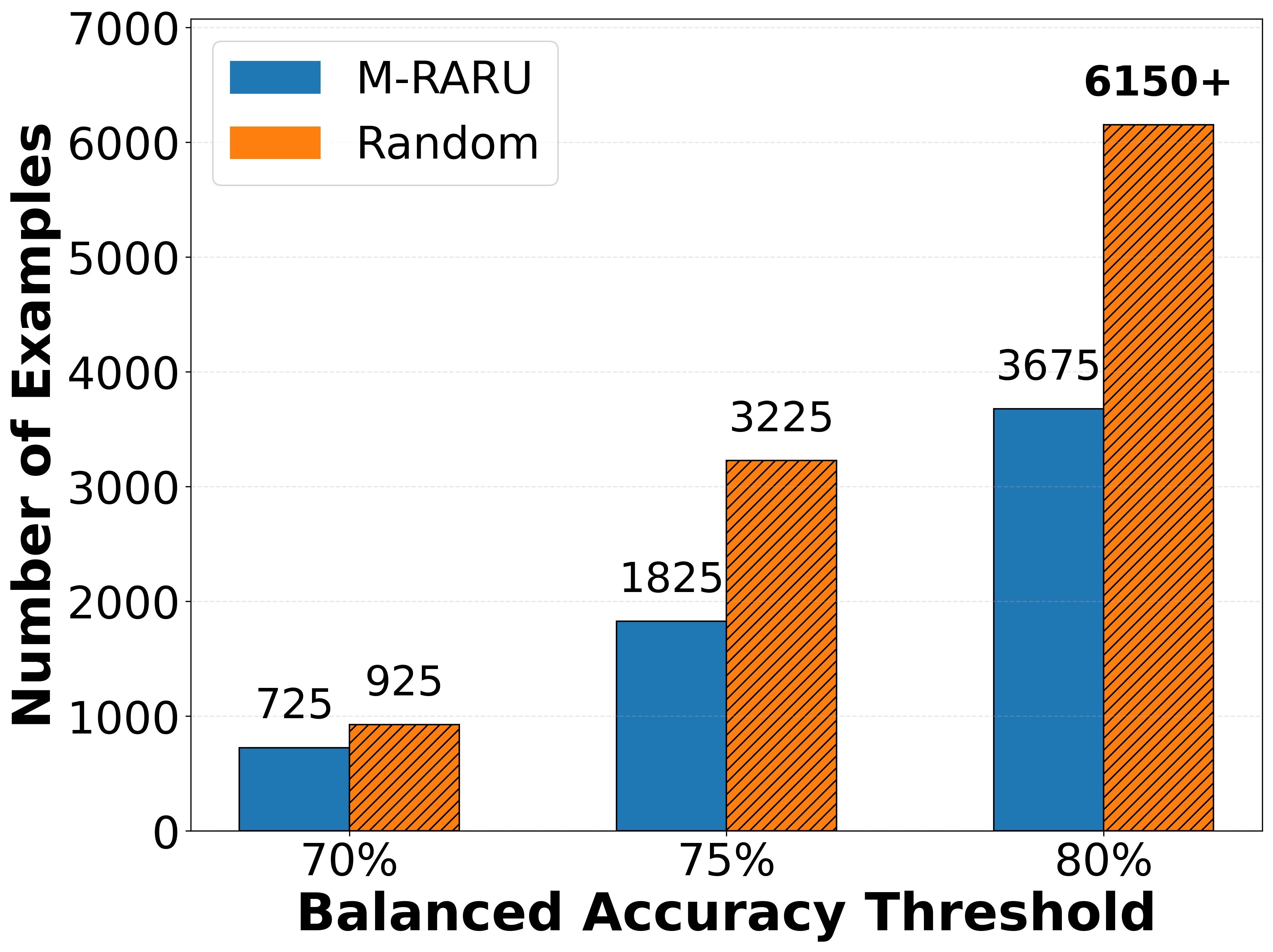}
        \caption{SVM GNAD Balanced Accuracy}
        \label{fig:svmtrnabal}
    \end{minipage}
\end{figure*}

%===============================================
% GNAD RESULTS - All Models
%===============================================

% --- Row 4: LDA GNAD (Acc & Bal Acc), RF GNAD (Acc & Bal Acc) ---
\begin{figure*}[!htb]
    \centering
    \begin{minipage}{0.24\textwidth}
        \centering
        \includegraphics[width=\linewidth]{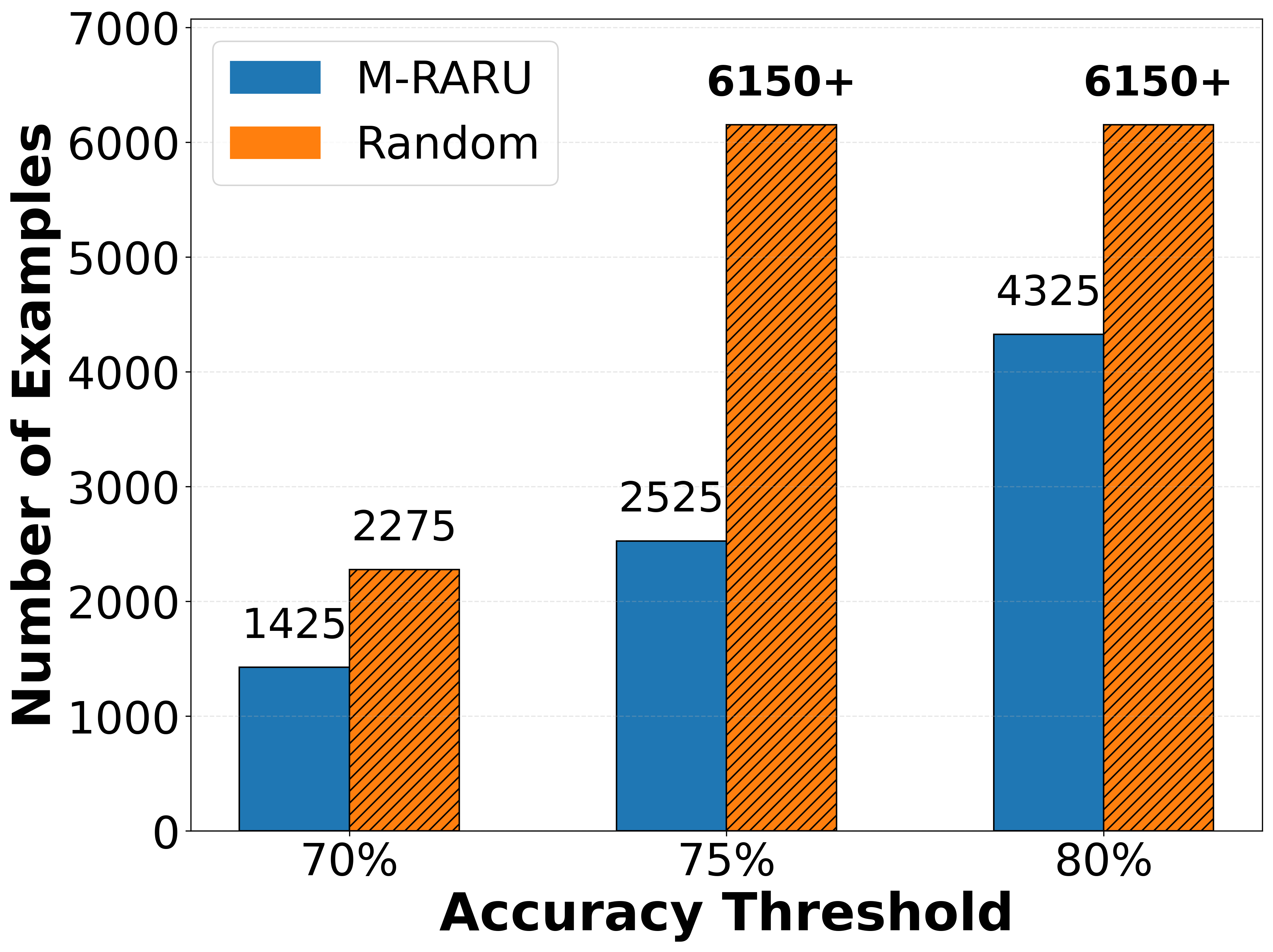}
        \caption{LDA GNAD Accuracy}
        \label{fig:ldatrnaacc}
    \end{minipage}\hfill
    \begin{minipage}{0.24\textwidth}
        \centering
        \includegraphics[width=\linewidth]{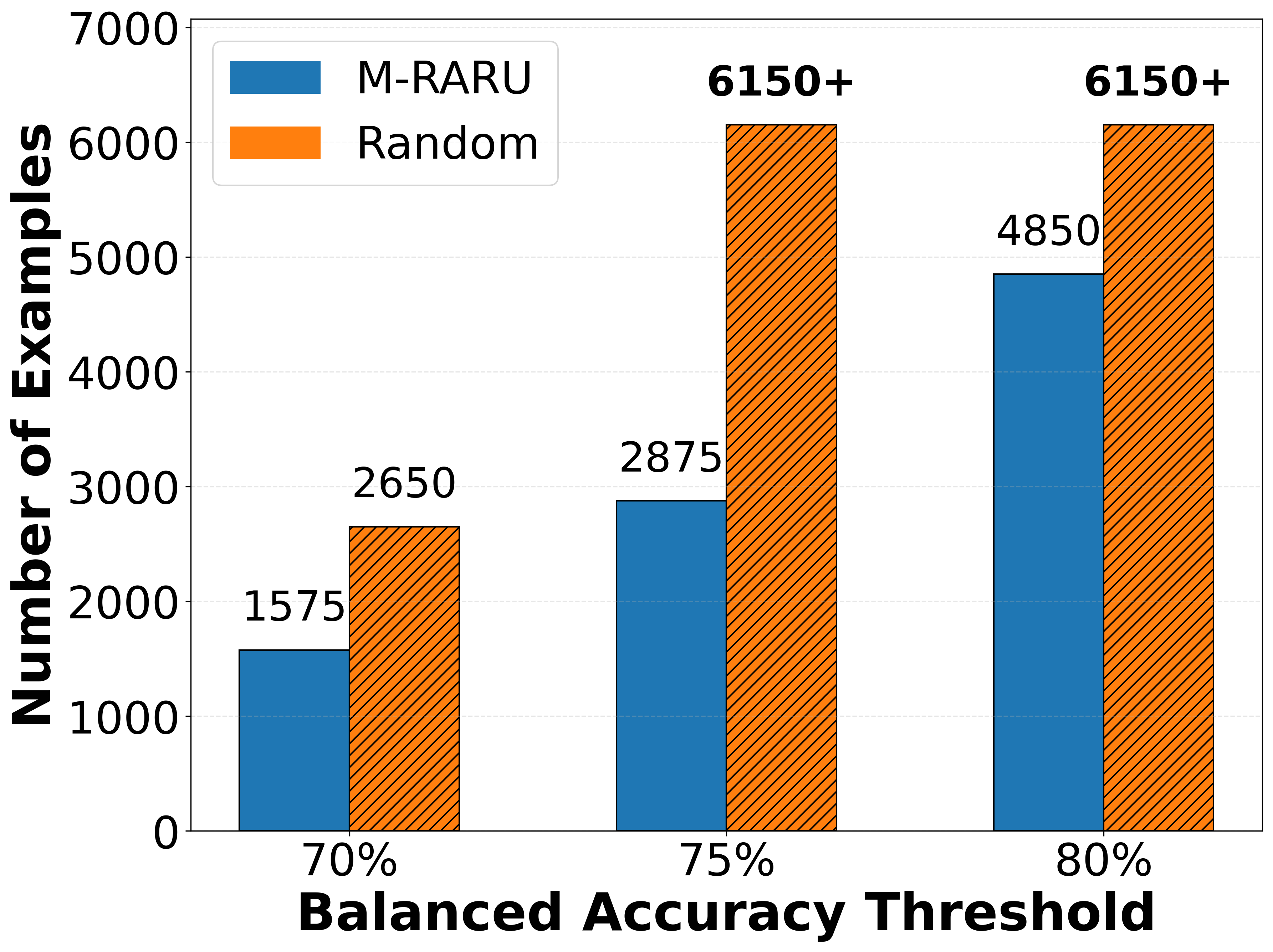}
        \caption{LDA GNAD Balanced Accuracy}
        \label{fig:ldatrnabal}
    \end{minipage}\hfill
    \begin{minipage}{0.24\textwidth}
        \centering
        \includegraphics[width=\linewidth]{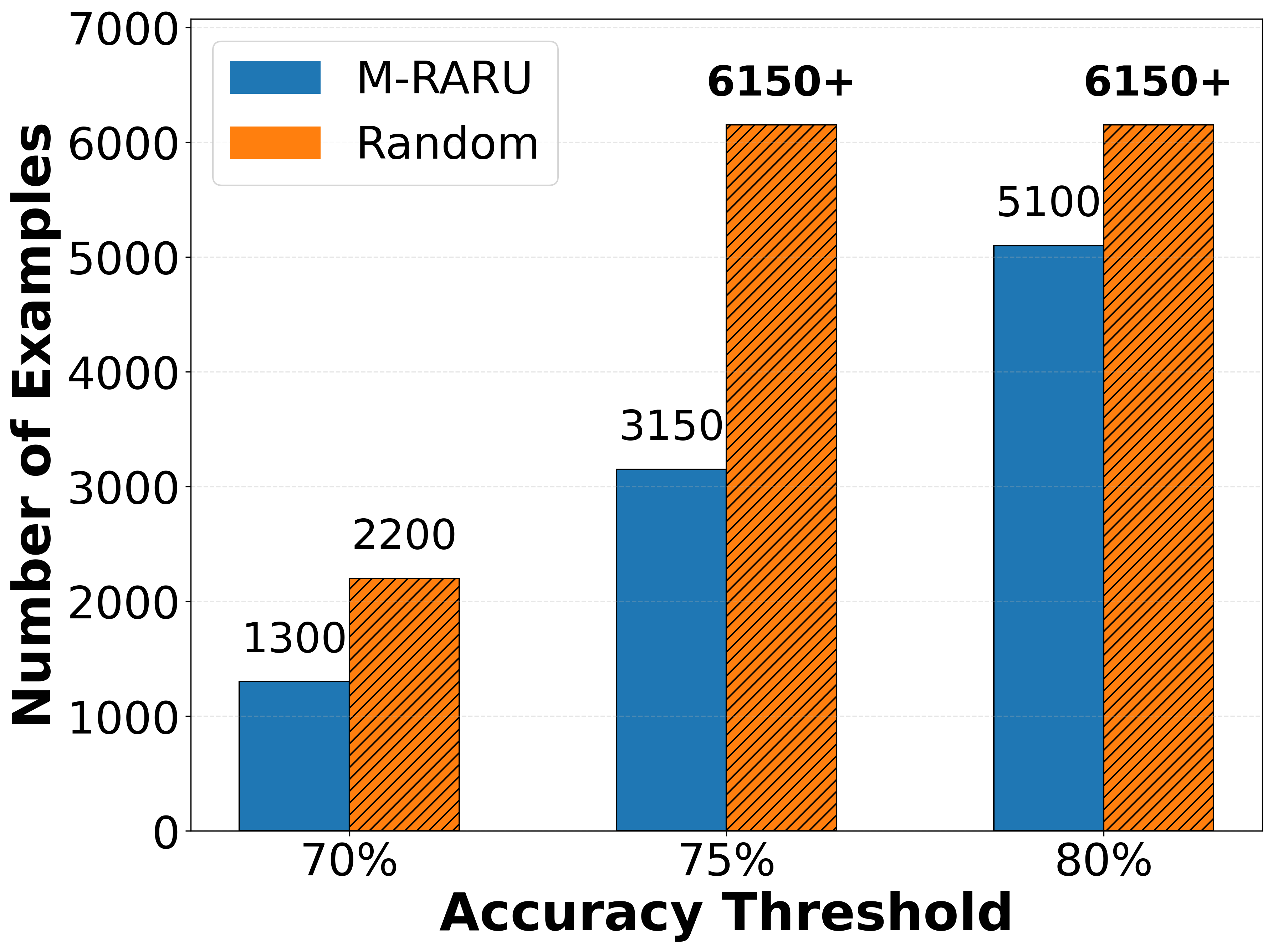}
        \caption{RF GNAD Accuracy}
        \label{fig:rftrnaacc}
    \end{minipage}\hfill
    \begin{minipage}{0.24\textwidth}
        \centering
        \includegraphics[width=\linewidth]{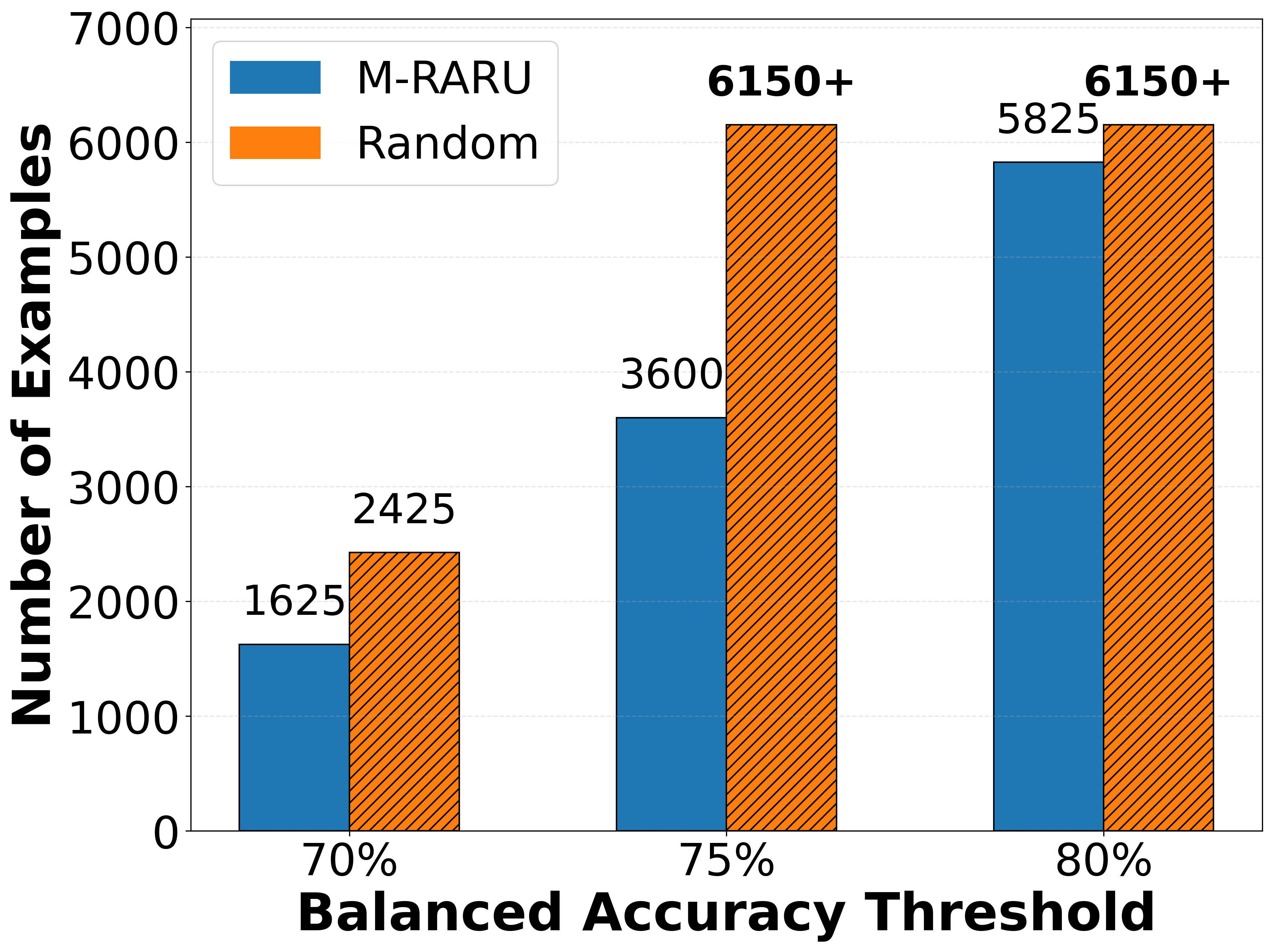}
        \caption{RF GNAD Balanced Accuracy}
        \label{fig:rftrnabal}
    \end{minipage}
\end{figure*}

% --- Row 5: GBDT GNAD (Acc & Bal Acc), DistilBERT GNAD (Acc & Bal Acc) ---
\begin{figure*}[!htb]
    \centering
    \begin{minipage}{0.24\textwidth}
        \centering
        \includegraphics[width=\linewidth]{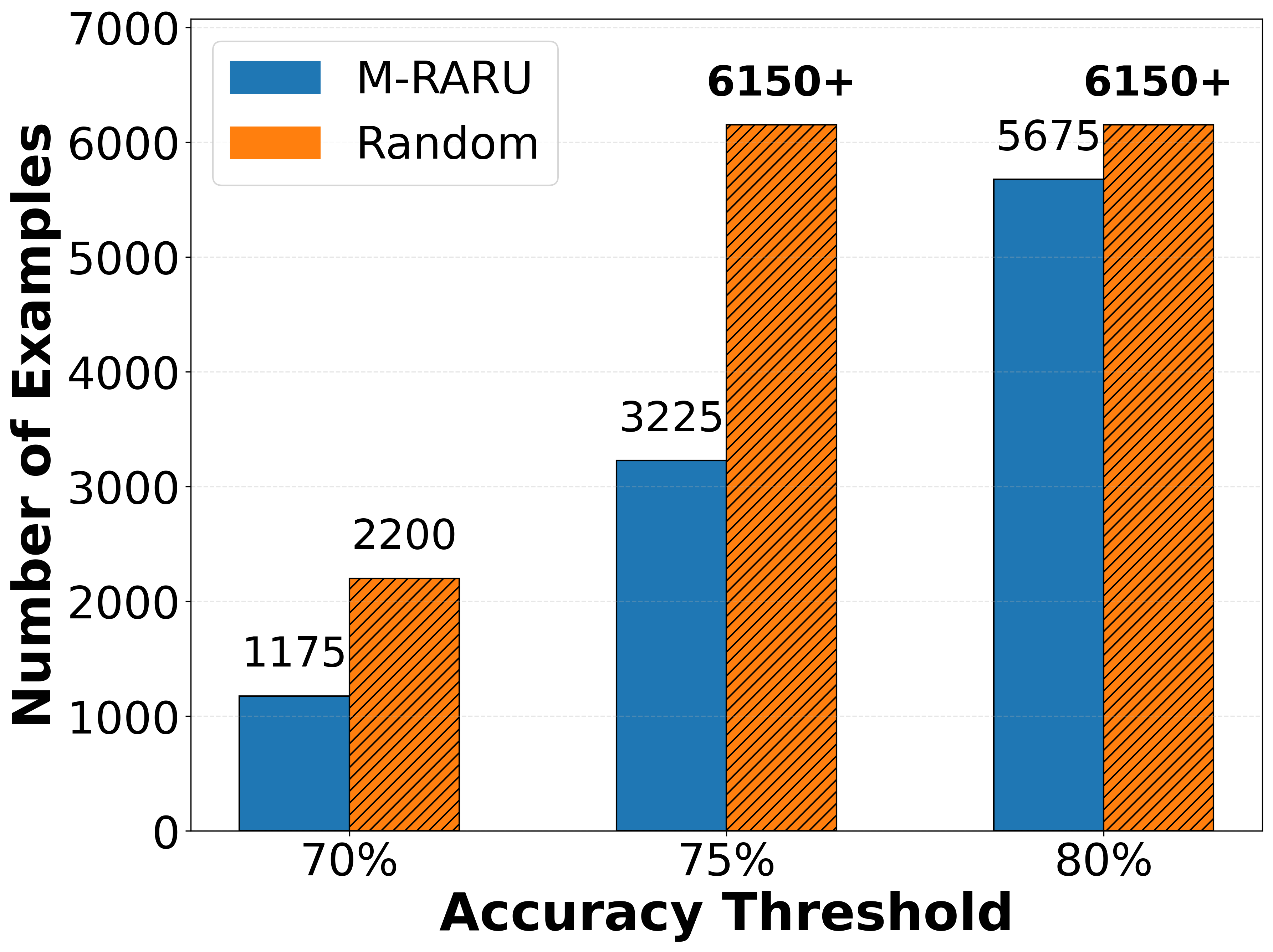}
        \caption{GBDT GNAD Accuracy}
        \label{fig:gbdttrnaacc}
    \end{minipage}\hfill
    \begin{minipage}{0.24\textwidth}
        \centering
        \includegraphics[width=\linewidth]{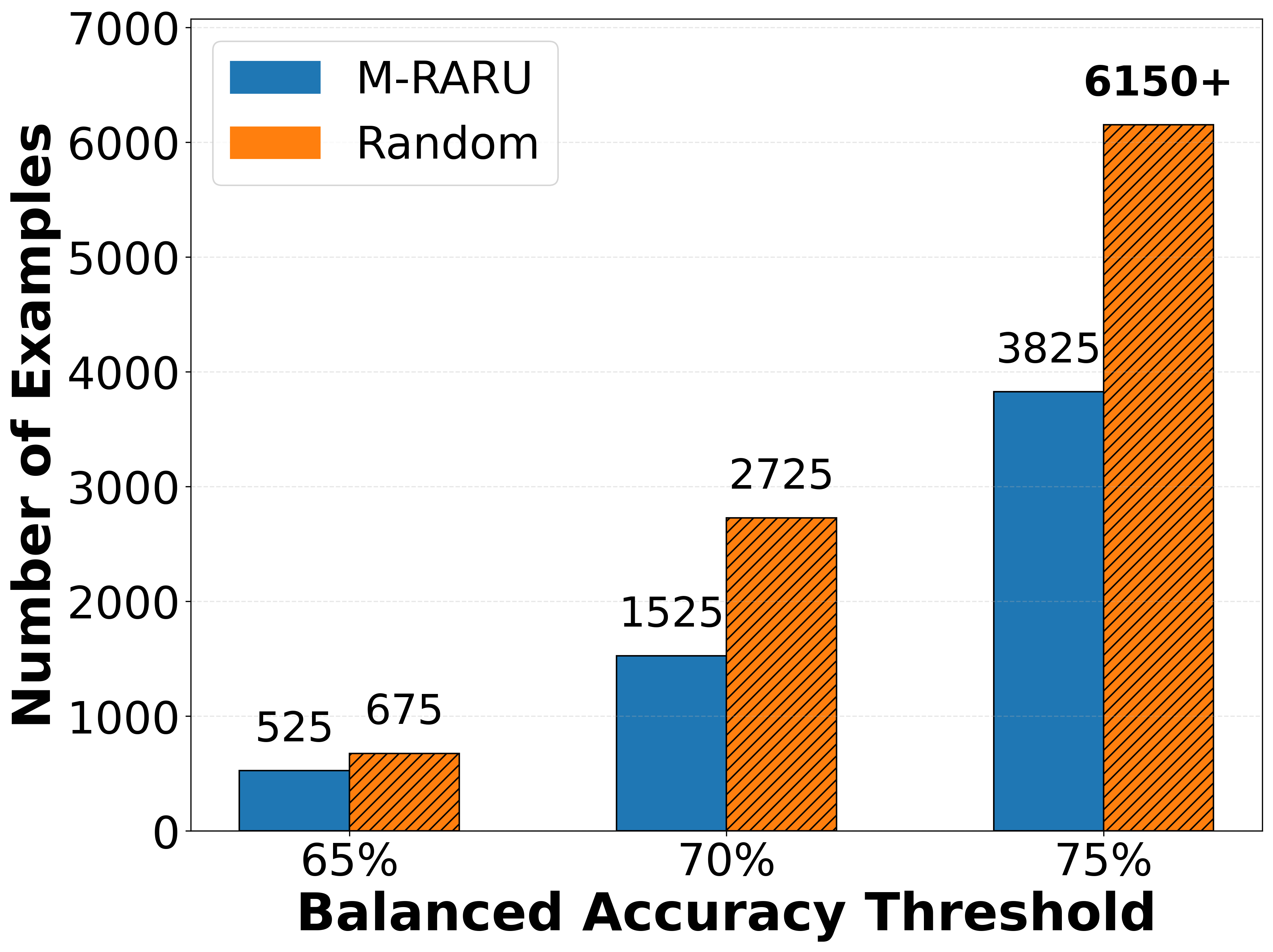}
        \caption{GBDT GNAD Balanced Accuracy}
        \label{fig:gbdttrnabal}
    \end{minipage}\hfill
    \begin{minipage}{0.24\textwidth}
        \centering
        \includegraphics[width=\linewidth]{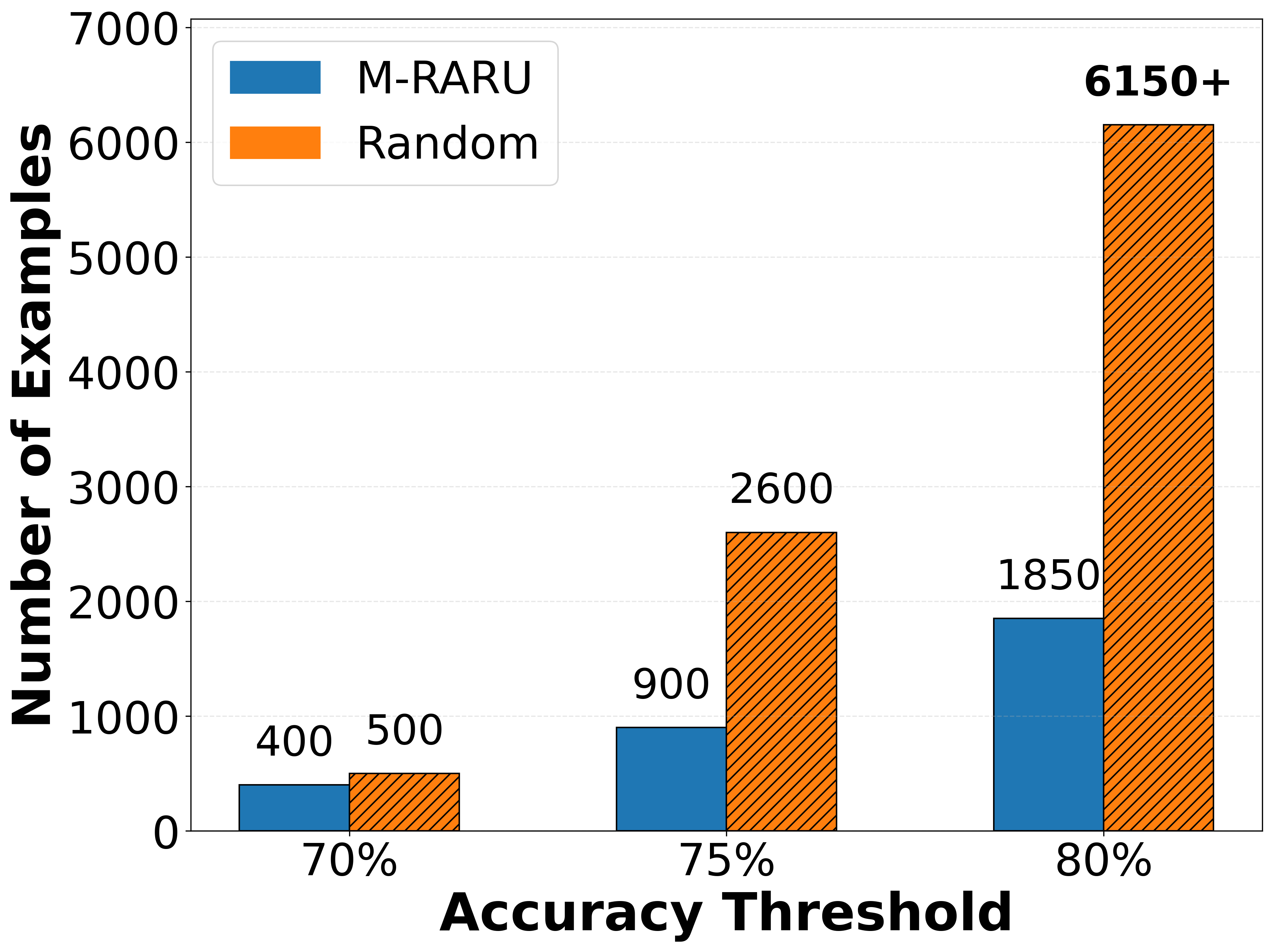}
        \caption{DistilBERT GNAD Accuracy}
        \label{fig:berttrnaacc}
    \end{minipage}\hfill
    \begin{minipage}{0.24\textwidth}
        \centering
        \includegraphics[width=\linewidth]{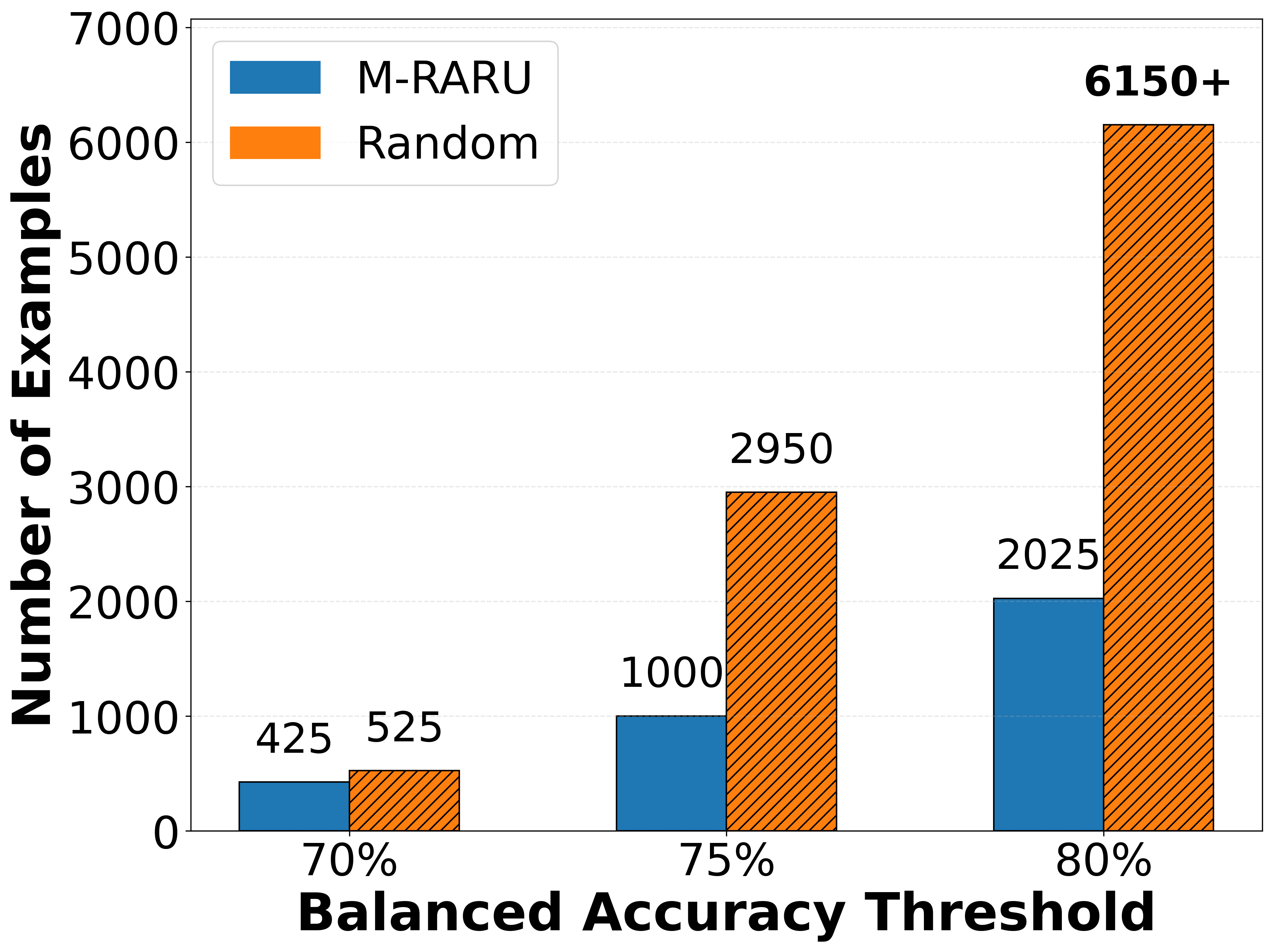}
        \caption{DistilBERT GNAD Balanced Accuracy}
        \label{fig:berttrnabal}
    \end{minipage}
\end{figure*}

\textbf{Datasets}
In our experiments, we used two real-world unstructured text datasets. \\
\textit{Public Comments Dataset:} This dataset comprises a vast collection of public responses to Federal Reserve announcements and regulations. For our experiments, we utilize a pool of 125,179 comments sampled from all public comments posted since 2008. The teacher model classifies each comment into one of five categories: Banks and Trades, Consumer/Community, Government, General Public, or Other, based on the commenter's organizational affiliation and perspective.\\
\textit{LSEG Data \& Analytics. Global News Archive Database (GNAD):} The GNAD dataset consists of professionally authored financial news articles. We utilize 12,288 news headlines for our experiments, focusing specifically on headline text to capture the most salient economic signals. The teacher model predicts whether each headline indicates rising, falling, or flat GDP trends, providing a concise economic sentiment classification task.

\textbf{Learning Representation}
To generate learning representations for the text, we employed the SentenceTransformer package. Specifically, we used the \texttt{all-MiniLM-L6-v2} model, which transforms each text segment into a 384-dimensional dense vector. These embeddings capture semantic relationships and serve as the unified feature space for any student models that requires an embedding (i.e., SVM, RF, GBDT, and LDA) in our experiments.

\textbf{Active Learning Schemes}
We experimented with one baseline scheme and our proposed scheme. In both schemes, selected examples are labeled by a teacher model, a locally deployed \texttt{gemma-3-27b-it-qat-q4\_0}, which acts as the oracle. The active learning process begins after an initial set of samples is randomly drawn to ensure at least one representative from each class is present in the training set.
\begin{itemize}
    \item \textbf{Random Sampling (RANDOM)}: The baseline scheme, where the system selects examples to be labeled from the unlabeled pool based on a uniform random distribution.
    \item \textbf{M-RARU}: Our proposed scheme, which uses Multi-class Randomized Accept/Reject Uncertainty Sampling to intelligently query the most informative examples for labeling by the teacher model.
\end{itemize}

\textbf{Student Models}
We evaluated our active learning schemes on five distinct student models to assess the generalizability of our approach. All traditional machine learning models use default scikit-learn configurations for training to ensure reproducibility and fair comparison. The models include: a Support Vector Machine (SVM), trained with default scikit-learn parameters; Linear Discriminant Analysis (LDA), using default scikit-learn configuration; a Random Forest (RF), an ensemble of decision trees with default scikit-learn settings; a Gradient-Boosting Decision Tree (GBDT), implemented using XGBoost for GPU support while maintaining default scikit-learn configuration parameters; and DistilBERT, a distilled version of BERT trained using default configurations from the Transformers library.

\textbf{Evaluation Metrics}
We assess the performance of the student models using two primary classification metrics.
\begin{enumerate}
    \item \textbf{Accuracy} is the proportion of correctly predicted instances over the total number of instances:
    \begin{equation}
    \text{Accuracy} = \frac{TP + TN}{TP + TN + FP + FN}
    \end{equation}
    \item \textbf{Balanced Accuracy} is the average of recall obtained on each class, which is suitable for imbalanced datasets:
    \begin{equation}
    \text{Balanced Accuracy} = \frac{1}{K} \sum_{i=1}^{K} \frac{TP_i}{TP_i + FN_i}
    \end{equation}
\end{enumerate}
where $TP$, $TN$, $FP$, and $FN$ are the counts of true positives, true negatives, false positives, and false negatives, respectively.

\textbf{Environment}
We implemented all algorithms in Python 3.11. All experiments were conducted on a machine equipped with a 16-core Intel CPU, 128GB of RAM, and a single NVIDIA V100 GPU with 32GB of memory. All reported results are averages of 5 complete runs, with the exception of the DistilBERT model, for which a single run was conducted due to computational constraints.

\textbf{Parameters}
Table \ref{t:params} provides a comprehensive list of the parameters and settings used throughout our experiments.

\subsection{Experimental Results}

\textbf{Accuracy Comparison}
Figures \ref{fig:svmpubacc} through \ref{fig:berttrnabal} present the primary results of our study, illustrating the performance of each student model under the M-RARU and RANDOM sampling schemes across both datasets. The y-axis of each plot represents either Accuracy or Balanced Accuracy, while the x-axis indicates the number of samples labeled by the teacher model. The accuracy thresholds shown in each model-dataset configuration represent the thresholds that are achievable with M-RARU within the given sample budget constraint (up to a cap of 90\%).

Our results demonstrate that M-RARU consistently outperforms RANDOM sampling across all model configurations, with the magnitude of improvement varying significantly based on the inherent uncertainty estimation capabilities of each model type. The variations in performance gains can be attributed to fundamental differences in how each model architecture estimates prediction uncertainty, which is a critical factor for active learning effectiveness.

\textbf{Tree-based Models (RF and GBDT)} exhibit the most dramatic yet inconsistent improvements with M-RARU. For instance, GBDT on Public Comments requires only 1,825 samples with M-RARU whereas RANDOM needs more than 6,275 to reach 90\% accuracy (71\% reduction in samples). However, there was little to no difference in necessary samples to reach the accuracy thresholds for RF. Tree-based models' native probabilistic outputs through ensemble voting mechanisms influences this performance. In Random Forests, the variance across individual tree predictions provides a naturally calibrated uncertainty estimate, while GBDT's sequential boosting process inherently focuses on difficult examples, aligning perfectly with M-RARU's uncertainty-driven selection. The discrete decision boundaries created by tree splits also produce clear regions of high uncertainty at class boundaries, making these models well-suited for identifying informative samples through active learning.

\textbf{Linear Models (SVM and LDA)} show substantial but more moderate improvements, typically achieving 50-70\% reductions in labeling requirements. To reach an accuracy of 90\% on the Public Comments data, LDA requires 2,250 samples using M-RARU compared to more than 6,275 with RANDOM (64\% reduction), and SVM requires only 875 samples using M-RARU compared to 2,075 using RANDOM (58\% reduction). These gains arise from the models' geometric interpretation of uncertainty. SVM's distance from the decision hyperplane provides a natural uncertainty metric that aligns well with M-RARU's sampling strategy, particularly effective in identifying support vectors that define class boundaries. LDA, as a generative model, offers well-calibrated posterior probabilities through its Gaussian assumptions, though its linear nature limits the complexity of uncertainty patterns it can capture compared to tree-based methods.

\textbf{DistilBERT} demonstrates the most modest improvements, with M-RARU typically requiring 10-20\% fewer samples than RANDOM. This limited benefit stems from several factors inherent to transformer architectures. First, as reported in \cite{DBLP:conf/emnlp/DesaiD20} DistilBERT's softmax outputs require additional calibration to produce reliable uncertainty estimates, as neural networks are known to be overconfident in their predictions. Second, the model's deep semantic understanding means it already performs well on randomly selected samples, reducing the relative benefit of strategic selection. Third, transformer models lack native uncertainty quantification mechanisms, unlike ensemble methods or Bayesian approaches, and require post-hoc techniques like temperature scaling or Monte Carlo dropout for uncertainty estimation. The computational overhead of these calibration methods further limits the practical benefits of active learning for transformer models.

\textbf{Dataset Complexity Impact.} The GNAD dataset consistently requires more samples across all models to achieve comparable accuracy levels, reflecting its more challenging classification task. News headlines, by nature, are extremely concise and often ambiguous, requiring sophisticated inference to determine GDP impact. Here, M-RARU's advantages become even more pronounced as many configurations with RANDOM sampling fail to reach higher accuracy thresholds within the 6,150 sample budget, while M-RARU is capable of achieving these targets. For example, Random Forest with RANDOM cannot reach 75\% accuracy on GNAD within the dataset limit, whereas M-RARU achieves this with only 2,200 examples.

\begin{table*}[t]
\caption{Student Model Inference and Training Comparison}
\centering
\small
\begin{tabular}{| c | c | c | c | c |}
\hline
\textbf{Model} & \textbf{Training (ms)} & \textbf{Training Speedup vs DistilBERT} & \textbf{Inference (ms)} & \textbf{Inference Speedup vs DistilBERT} \\
\hline
DistilBERT (CUDA) & 13.3 & 1.0× & 2.80 & 1.0× \\
\hline
GBDT (CUDA) & 0.3 & 44× & 0.08 & 35× \\
\hline
RF & 1.5 & 9× & 0.12 & 23× \\
\hline
LDA & 4.9 & 3× & 0.22 & 13× \\
\hline
SVM & 1.5 & 9× & 0.55 & 5× \\
\hline
\end{tabular}
\label{table:model_performance_speedup}
\end{table*}

\textbf{Balanced Accuracy Analysis.} When examining balanced accuracy metrics, which better account for class imbalance, the benefits of M-RARU become even more apparent. The strategic sampling inherently addresses class imbalance by focusing on decision boundaries where minority classes are often found. For instance, RF on Public Comments requires over 6,275 examples with RANDOM to achieve 80\% balanced accuracy, while M-RARU needs only 1,200, representing an 81\% reduction. This improvement is particularly valuable in real-world applications where minority classes often represent critical but rare events.

The consistent pattern across all experiments reveals that M-RARU's effectiveness scales with model uncertainty quality: models with naturally calibrated uncertainties (tree ensembles) benefit most, followed by models with geometric uncertainty interpretations (SVM, LDA), while models requiring uncertainty calibration (DistilBERT) show modest but still meaningful improvements. These results validate our hypothesis that combining knowledge distillation with intelligent active learning can dramatically reduce the cost of creating high-performance classifiers.

\begin{table}[t]
\caption{Sampling Efficiency: M-RARU vs Traditional Uncertainty Sampling}
\centering
\small
\begin{tabular}{| c | c | c | c | c |}
\hline
\textbf{Model} & \multicolumn{2}{c|}{\textbf{Public Comments}} & \multicolumn{2}{c|}{\textbf{GNAD}} \\
\cline{2-5}
 & \textbf{Acc. Rate} & \textbf{Speedup} & \textbf{Acc. Rate} & \textbf{Speedup} \\
\hline
SVM & 18.2\% & 912× & 31.3\% & 154× \\
\hline
LDA & 0.2\% & 10× & 1.9\% & 9× \\
\hline
RF & 35.7\% & 1,788× & 42.9\% & 211× \\
\hline
GBDT & 13.7\% & 686× & 33.1\% & 163× \\
\hline
DistilBERT & 5.9\% & 295× & 8.3\% & 41× \\
\hline
\end{tabular}
\label{table:mraru_efficiency}
\end{table}

\subsection{Training Efficiency Analysis}
Table \ref{table:model_performance_speedup} illustrates the computational efficiency gains achieved by traditional machine learning models compared to the transformer-based DistilBERT baseline. These measurements represent averages across 1,000 batches of 32 samples each.

In terms of training efficiency, GBDT demonstrates exceptional performance with a 44x speedup compared to DistilBERT, requiring only 0.3ms per batch versus 13.3ms for the transformer model. This dramatic improvement stems from GBDT's sequential tree construction algorithm, which efficiently leverages gradient information without the computational overhead of backpropagation through deep neural networks. Random Forest and SVM both achieve 9x training speedups, completing batch training in 1.5ms through parallelizable training procedures. LDA shows a 3x speedup with 4.9ms training time, as its statistical approach requires matrix operations that, while efficient, are more computationally intensive than tree-based methods.

For inference performance, the advantages become even more pronounced. GBDT achieves a 35x speedup with inference times of 0.08ms per batch, making it ideal for real-time applications. Random Forest delivers 23x faster inference at 0.12ms through simple tree traversal operations, while LDA provides 13x faster inference at 0.22ms via straightforward linear transformations.

These efficiency gains have profound implications for model development. The time saved by faster models can be directly reinvested into hyperparameter tuning, which is a critical process for maximizing predictive performance \cite{bergstra2012random, snoek2012practical}. Within a fixed time budget, a practitioner can execute hundreds of GBDT experiments in the time required for a single DistilBERT run. This enables thorough exploration of the hyperparameter space, dramatically increasing the probability of finding optimal configurations.

The combination of M-RARU's sample efficiency and traditional models' computational speed creates a multiplicative advantage: M-RARU reduces labeling time while efficient models accelerate training, enabling rapid iteration cycles. For instance, within a single workday, one could test hundreds of combinations of learning rates, tree depths, and regularization parameters for GBDT. The same search would take weeks with transformer models. This capability ensures that knowledge distilled from the teacher LLM is leveraged to its fullest extent, producing models that are not only fast but optimally tuned for peak performance.

\subsection{Sampling Efficiency Analysis}
Previously, \cite{request} and \cite{exnav} have shown that the randomized accept/reject mechanism achieves comparable performance to the traditional exhaustive-based uncertainty sampling. To further strengthen the comparison, in Table \ref{table:mraru_efficiency}, we quantify the computational efficiency gains of M-RARU over traditional uncertainty sampling when reaching 85\% accuracy for Public Comments and 75\% accuracy for GNAD. The calculations assume a batch size of 25 samples, where traditional uncertainty sampling must perform exhaustive searches through the entire unlabeled pool (125,179 samples for Public Comments, 12,288 for GNAD) after training each batch to identify the most uncertain samples. In contrast, M-RARU employs the accept/reject mechanism described in Equation \ref{raruaccept}, where the uncertainty score directly serves as the acceptance probability, eliminating the need for exhaustive ranking. The acceptance rates shown reflect the average probability of accepting a sample during the active learning process until these accuracy thresholds are reached. The results reveal striking variations in acceptance rates across models: tree-based methods (RF and GBDT) maintain healthy acceptance rates of 13.7-42.9\%, yielding speedups of 163-1,788× when reaching target accuracy, while SVM shows intermediate rates of 18.2-31.3\% with speedups of 154-912×. Most notably, LDA exhibits pathologically low acceptance rates of 0.2\% on Public Comments and 1.9\% on GNAD, resulting in minimal speedups of 10× and 9× respectively. This poor performance stems from LDA's generative modeling approach, which produces overly confident posterior probabilities concentrated in narrow regions of the feature space. When LDA assigns high confidence to most samples (leaving few truly uncertain), the acceptance probability $p = 1 - \max_k \Pr(C_k|\mathbf{x})$ becomes vanishingly small for the vast majority of the pool, and thus, leads to more candidate being exam. Overall, as can be seen from the results, thanks to the adaption of randomized accept/reject mechanism, M-RARU is requiring far less inferences than any traditional active learning samplings that requires exhaustive search.
\section{Related Works}
\label{sec:rw}

In this section, we will present the works that are closely related to our research. We begin by introducing the literature on Knowledge Distillation, with a particular focus on its application to text classification. Then, we discuss established principles in Active Learning for efficient data selection. Finally, we survey the emerging intersection of these two fields, which provides the context for our proposed methodology.

\subsection*{Knowledge Distillation for Text Classification}
The concept of Knowledge Distillation (KD) was formally introduced as a method to compress large, complex models into smaller, more efficient ones without a significant loss in performance \cite{hinton2015distillingknowledgeneuralnetwork}. The fundamental idea is to train a compact "student" model to mimic the behavior of a larger, pre-trained "teacher" model. This is typically achieved by using the softened class probabilities produced by the teacher as soft labels to guide the student's training process.
In the domain of Natural Language Processing (NLP), this technique gained significant traction with the advent of large-scale transformer models. For instance, works like DistilBERT \cite{sanh2019distilbert} and TinyBERT \cite{jiao2019tinybert} demonstrated that it was possible to create much smaller and faster versions of BERT that retained over 95\% of the original model's performance on standard NLP benchmarks.
Specifically for text classification, KD has been explored in various contexts. Some approaches focus on distilling knowledge across different domains, training a student model for a target domain using teachers with expertise in related source domains \cite{ZHANG202211}. Others have adapted KD for industrial applications, developing performance-guided strategies to create efficient classifiers at scale by carefully selecting the knowledge to be transferred \cite{Palo2024}. There is also research on distilling knowledge between different modalities, such as from text-based models to speech-based models \cite{ni2023adaptiveknowledgedistillationtext}. Despite these advancements, a common challenge in nearly all KD applications is the high cost associated with the initial step: requiring the powerful but slow and expensive teacher model to label a very large, randomly sampled dataset to create the training set for the student \cite{yuan2021revisiting}.

\subsection*{Active Learning for Efficient Model Training}
Active Learning (AL) is a subfield of machine learning that aims to reduce the total amount of labeled data required to train a model by allowing the learning algorithm to intelligently choose the data from which it learns \cite{Settles10activelearning}. The core principle is that not all data points are equally informative. By iteratively selecting the most valuable samples for labeling, an AL system can achieve a desired level of performance with significantly fewer labels than required by passive, random sampling approaches.
A wide variety of query strategies have been developed to identify informative samples. The most common approach is uncertainty sampling, where the algorithm queries the instances about which it is least certain of the correct label \cite{Lewis94a}. Other popular strategies include Query-by-Committee (QBC), which uses an ensemble of models and selects samples on which the committee members disagree the most \cite{COLT::SeungOS1992}, and Expected Model Output Change (EMOC), which prioritizes samples that are expected to cause the greatest change to the current model if their labels were known \cite{freytag14sie}.
Another related technique is importance sampling, which has a rich history in statistics. In machine learning, it is used to prioritize data points that have a larger impact on the model's loss function, thereby reducing training time and improving final accuracy \cite{katharopoulos2019samplescreatedequaldeep}. Recent work has extended this to create task-adaptive pretraining schemes by sampling data that is most relevant to the target task \cite{grangier2025taskadaptivepretrainedlanguagemodels}. Our work draws inspiration from these principles, but applies them to the unique problem of cost-effective knowledge transfer from a teacher model.
More recently, \cite{request} and \cite{exnav} introduced a Randomized Accept/Reject mechanism into Uncertainty Sampling, which addresses the scalability issues of traditional uncertainty sampling through probabilistic selection. However, their implementations were limited to binary classification tasks, and thus, are unsuited to the particular objective of this work.

\subsection*{Integrating Active Learning and Knowledge Distillation}
The high cost of data annotation in standard KD has naturally led researchers to explore the integration of AL. The goal of this hybrid approach, often termed Active Knowledge Distillation (AKD), is to use AL query strategies to select a small, highly informative subset of unlabeled data for the teacher LLM to label, thereby minimizing expensive API calls and computational overhead \cite{wang2023active, zhang2023active}.
Several strategies have been proposed within this emerging area. Some methods use traditional uncertainty metrics, where the student model identifies confusing samples and requests teacher labels only for those \cite{kothadiya2022task}. Others have developed more sophisticated metrics that consider both the student's uncertainty and the teacher's confidence, aiming to select samples that are not only hard for the student but also confidently labeled by the teacher \cite{kuznetsov2021active}. Furthermore, research has shown that co-training frameworks, where the student and teacher models are trained simultaneously in an active learning loop, can yield more robust results \cite{xie2020active}. A comprehensive survey of data selection methods highlights the critical role that strategic sampling plays in the overall efficiency of training modern language models \cite{trainingdatamethods}.
However, many existing AKD methods still rely on deterministic uncertainty sampling, which can be prone to selecting outliers and may not sufficiently explore the data space. These methods often lack a mechanism to balance exploration (sampling from diverse regions) and exploitation (sampling from regions of high uncertainty). Our proposed algorithm, M-RARU, addresses this specific gap. It integrates a randomized accept-reject mechanism with uncertainty sampling, providing a principled way to manage the exploration-exploitation trade-off and cost-effectively select a diverse and highly informative training dataset for the student model. 
\section{Conclusion}
\label{sec:con}
In this work, we study the problem of cost-effective model training for large-scale text classification. To address this, we proposed a novel approach that combines Knowledge Distillation with Active Learning for efficient knowledge transfer.
This approach effectively transfers a Large Language Model teacher's knowledge to a smaller student model, creating highly accurate classifiers that achieve a level of performance difficult to obtain with traditional training methods alone.
Our proposed method enables knowledge transfer for any student model as long as it can provide a measure of predictive uncertainty.
In addition, we described in detail the key component of this approach, namely, Multi-class Randomized Accept/Reject Uncertainty Sampling (M-RARU), an intelligent query strategy that optimizes the selection of training instances for the LLM teacher.
We implemented our approach and experimentally verified its performance with five distinct student models on multiple real-world datasets. The results have shown that our proposed method exhibits substantially better performance when compared to the random sampling baseline while achieving desired classification accuracy. Specifically, M-RARU achieves up to 80\% reduction in sample requirements compared to random sampling, substantially reducing the required training data and associated labeling costs while achieving the same, or greater, accuracy as the baseline alternative.
\smallskip
%\noindent\textbf{Acknowledgment:} We would like to thank... 

\begin{small}
\begin{spacing}{0.94}
\bibliographystyle{IEEEtran}
\bibliography{all_ref}
\end{spacing}
\end{small}

\end{document}